\documentclass[letterpaper]{article} 
\usepackage{aaai24}  
\usepackage{times}  
\usepackage{helvet}  
\usepackage{courier}  
\usepackage[hyphens]{url}  
\usepackage{graphicx} 
\usepackage{natbib}  
\usepackage{caption} 
\usepackage{algorithm}
\usepackage{algorithmic}

\usepackage{newfloat}
\usepackage{listings}
\DeclareCaptionStyle{ruled}{labelfont=normalfont,labelsep=colon,strut=off} 
\floatstyle{ruled}
\newfloat{listing}{tb}{lst}{}
\floatname{listing}{Listing}
\usepackage{multirow}

\usepackage{booktabs}       
\usepackage{amsfonts}       
\usepackage{amsthm}
\usepackage{amsmath}

\usepackage{tikz}
\usetikzlibrary{bayesnet}
\usetikzlibrary{arrows.meta}
\usepackage{subfigure}

\tikzstyle{causallatent} = [circle,fill=white,draw=black,inner sep=1pt,
minimum size=22pt, font=\fontsize{10}{10}\selectfont, node distance=1]
\tikzstyle{causalobs} = [causallatent,fill=gray!25]

\tikzstyle{causaleps} = [rectangle,fill=white,draw=black,inner sep=1pt,
minimum size=18pt, font=\fontsize{8}{8}\selectfont, node distance=1]

\tikzstyle{causalobsset} = [ellipse,fill=white,draw=black,inner sep=1pt,
minimum size=18pt, font=\fontsize{8}{8}\selectfont, node distance=1, fill=gray!25]

\renewcommand{\edge}[3][]{ %
  \foreach \x in {#2} { %
    \foreach \y in {#3} { %
      \path (\x) edge [-To, #1] (\y) ;%
    } ;
  } ;
}
\theoremstyle{plain}
\newtheorem{theorem}{Theorem}[section]

\theoremstyle{definition}
\newtheorem{definition}[theorem]{Definition}

\theoremstyle{remark}

\title{Identification of Causal Structure with Latent Variables Based on Higher Order Cumulants}
\author {
    Wei Chen\textsuperscript{\rm 1},
    Zhiyi Huang\textsuperscript{\rm 1},
    Ruichu Cai\textsuperscript{\rm 1,\rm 2}\thanks{Corresponding author.},
    Zhifeng Hao\textsuperscript{\rm 1,\rm 3},
    Kun Zhang\textsuperscript{\rm 4,\rm 5}
}
\affiliations {
    \textsuperscript{\rm 1} School of Computer Science, Guangdong University of Technology, Guangzhou, China\\
    \textsuperscript{\rm 2} Peng Cheng Laboratory, Shenzhen, China\\
    \textsuperscript{\rm 3} College of Engineering, Shantou University, Shantou, China\\
    \textsuperscript{\rm 4} Department of Philosophy, Carnegie Mellon University, Pittsburgh, PA, United States\\
    \textsuperscript{\rm 5} Mohamed bin Zayed University of Artificial Intelligence, Abu Dhabi, United Arab Emirates\\
    \{chenweidelight, huangzhiyichn, cairuichu\}@gmail.com, zfhao@gdut.edu.cn, kunz1@cmu.edu
}

\usepackage{bibentry}

\begin{document}

\maketitle

\begin{abstract}
Causal discovery with latent variables is a crucial but challenging task. Despite the emergence of numerous methods aimed at addressing this challenge, they are not fully identified to the structure that two observed variables are influenced by one latent variable and there might be a directed edge in between. Interestingly, we notice that this structure can be identified through the utilization of higher-order cumulants. By leveraging the higher-order cumulants of non-Gaussian data, we provide an analytical solution for estimating the causal coefficients or their ratios. With the estimated (ratios of) causal coefficients, we propose a novel approach to identify the existence of a causal edge between two observed variables subject to latent variable influence. In case when such a causal edge exits, we introduce an asymmetry criterion to determine the causal direction. The experimental results demonstrate the effectiveness of our proposed method.
\end{abstract}

\section{Introduction}

\begin{figure}[t]
    \centering
    \subfigure[]{
        \begin{tikzpicture}[thick]
            \node[causallatent] (L) {$L$}; %
            \node[causalobs, xshift= -25pt, yshift=-40pt] (X) {$X$} ;
            \node[causalobs, xshift=  25pt, yshift=-40pt] (Y) {$Y$} ;
            \edge{L}{X, Y};
            \draw (-20pt, -18pt) node [inner sep=0.75pt]    {$\lambda_{1}$};
            \draw ( 20pt, -18pt) node [inner sep=0.75pt]    {$\lambda_{2}$};
        \end{tikzpicture}
    }    
    \hspace{5pt}
    \subfigure[]{
        \begin{tikzpicture}[thick]
            \node[causallatent] (L) {$L$}; %
            \node[causalobs, xshift= -25pt, yshift=-40pt] (X) {$X$} ;
            \node[causalobs, xshift=  25pt, yshift=-40pt] (Y) {$Y$} ;
            \edge{L}{X, Y};
            \draw (-20pt, -18pt) node [inner sep=0.75pt]    {$\lambda_{1}$};
            \draw ( 20pt, -18pt) node [inner sep=0.75pt]    {$\lambda_{2}$};
            \edge{X}{Y};
            \draw (  0pt, -36pt) node [inner sep=0.75pt]    {$\eta$};
        \end{tikzpicture}
    }    
    \caption{A causal structure with two observed variables $X$ and $Y$ affected by one latent variable $L$, where $\lambda_1$, $\lambda_2$ and $\eta$ denote the causal strength between $L$ and $X$, between $L$ and $Y$, between $X$ and $Y$, respectively.}
    \label{fig:example}
\end{figure}
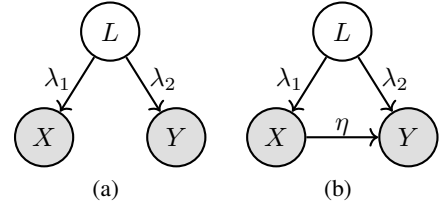

Inferring causal relationships between observed variables with latent variables is of significant importance and has been applied in many fields \cite{sachs2005causal,wang2020high,tramontano2022learning,morioka2023connectivity}. 
The Latent Variables Linear Non-Gaussian Acyclic Model (LvLiNGAM) \cite{hoyer2008estimation,entner2011discovering, tashiro2014parcelingam} is one of the most prominent approaches for this problem. Notably, LvLiNGAM can be transformed into a canonical model, wherein each latent variable is the cause of a minimum of two children and has no parents \cite{hoyer2008estimation}. Therefore, the identification of the structure involving one latent variable and two observed variables (as shown in Figure \ref{fig:example}) is one of the fundamental problems in causal discovery.

Various methods are proposed to identify the causal structure with latent variables. One of the typical methods is based on conditional independence tests, such as FCI \cite{spirtes1995causal} and its variants \cite{colombo2014learning}, but they can only identify up to Markov equivalent class. Under the measurement assumption, some approaches use rank constraints \cite{silva2006learning,kummerfeld2016causal,huang2022latent}, and Generalized Independence Noise (GIN) condition \cite{xie2020generalized,xie2022identification} to identify the relationships between latent variables and observed variables. By utilizing the non-Gaussianity, PraceLiNGAM \cite{tashiro2014parcelingam}, MLCLiNGAM \cite{chen2021causal} and RCD \cite{maeda2020rcd} can only identify some causal structure among observed variables that are not directly affected by the latent variables. However, the above approaches fail to identify the structures given in Figure \ref{fig:example}. 
Although lvLiNGAM leverages the Overcomplete Independence Component Analysis technique (ICA) \citep{eriksson2004identifiability, lewicki2000learning}, it is hard to obtain an optimal result, which would lead to wrong causal relations.
In summary, how to fully identify the causal relationships between two observed variables influenced by one latent variable is still an open problem. 


To overcome the aforementioned challenge, two problems should be considered: 1) How to detect whether there exists a causal edge between two observed variables? 2) Furthermore, when such an edge exists, how to determine its direction? 
Fortunately, we notice that certain types of non-Gaussianity can be measured by higher-order cumulants, which can also capture some data distribution information \citep{aapo2001independent}. Thus, utilizing higher-order cumulants can yield additional information to address the aforementioned issues. 

To address the first problem, we find that some specific combinations of higher-order joint cumulants of two observed variables can be used to detect whether there exists a causal edge between them. In Figure \ref{fig:example} (a), the joint cumulants of $X$ and $Y$ can be approximately regarded as the multiplication of powers of coefficients and the cumulants of their shared components, which can be expressed as follows:
\begin{equation}
    \mathcal{C}_{i,j}(X, Y) = \lambda_1^{i} \lambda_2^{j} \mathcal{C}_{i+j}(L), i,j>0,
\end{equation} 
where $\mathcal{C}_{i, j}(X, Y)$ represents the joint cumulant $Cum(\underbrace{X, \dots}_{i \,\text{times}}, \underbrace{Y, \dots}_{j \,\text{times}})$. This distinction arises from the fact that $X$ and $Y$ only share one latent component $L$ in Figure \ref{fig:example}(a), whereas they share two components $L$ and the noise of $X$ in Figure \ref{fig:example}(b). Then, asymptotically, if and only if no causal directed edge exists between $X$ and $Y$, the following constraint holds:
\begin{equation}
    \mathcal{C}_{i+1, j+1}(X, Y)^2 = \mathcal{C}_{i, j+2}(X, Y) \mathcal{C}_{i+2, j}(X, Y).
    \label{eq:no_edge}
\end{equation}
In such a case, we can also identify the causal effects of the single latent variable on $X$ and $Y$ by leveraging the higher-order cumulants on the non-Gausian data.

The next problem is how to determine the causal direction between two observed variables. Interestingly, considering data related to observed variables $X$ and $Y$ generated according to Figure \ref{fig:example}(b), leveraging the third-order cumulants of $X$ and $Y$, one have
\begin{equation}
    \begin{aligned}
        \mathcal{C}_3(X) & = \lambda_1^3\mathcal{C}_3(L)+\mathcal{C}_3(E_X),\\
        \mathcal{C}_3(Y) & \neq (\lambda_1 \eta+\lambda_2)^3\mathcal{C}_3(L) + \eta^3 \mathcal{C}_3(E_X),
    \end{aligned}
\end{equation}
where $E_X$ and $E_Y$ represent noise terms associated with $X$ and $Y$ respectively, and $L$ denotes a latent variable. This discrepancy arises from the presence of an additional term $E_Y$ in $Y$, which is distinct from $X$. Thus, we can develop an asymmetry criterion to determine the causal direction between two variables in the presence of latent variables.

\section{Preliminary}

\subsection{Latent-Variable Linear Non-Gaussian Acyclic Model}
In this paper, we consider the data over observed variables $\mathbf{V}$ that may be affected by the latent variables $\mathbf{L}$. Specifically, these data are generated from a linear causal model with non-Gaussian noises, which can be formalized as:

\begin{equation}
\mathbf{V}=\mathbf{B}\mathbf{V}+\mathbf{\Lambda}\mathbf{L}+\mathbf{E},
\label{eq:model}
\end{equation}
where $\mathbf{B}$ is the causal strength matrix, $\mathbf{\Lambda}$ is the causal strength matrix from $\mathbf{L}$ to $\mathbf{V}$, $\mathbf{E}$ is a non-Gaussian noise term and each $E_i \in \mathbf{E}$ is independent of others. This model is also termed Latent-Variable Linear Non-Gaussian Acyclic Model (LvLiNGAM) \cite{hoyer2008estimation}.

In the linear case, many researchers are likely to transform the above model into OICA model \cite{lewicki2000learning,eriksson2004identifiability}, $\mathbf{V} = \mathbf{A}\mathbf{S}$, where $\mathbf{A}$ is represented as the mixing matrix and $\mathbf{S}$ denotes the latent independent components. For example, in Figure \ref{fig:example}(b), according to the data generation process described by Eq. (\ref{eq:model}), $X$ and $Y$ can be formalized as:
\begin{equation}
    \begin{aligned}
    X & = \lambda_{1}L + E_X = \alpha_{1}L + \beta_{1}E_X,\\
    Y & = \lambda_{2}L + \eta X + E_Y = \alpha_{2}L + \beta_{2}E_X + \gamma_{2}E_Y,
    \end{aligned}
    \label{eq:figure2}
\end{equation}
where $\lambda_{1}$ and $\lambda_{2}$ are the causal strengths from $L$ to $X$ and $Y$, respectively; $\eta$ represents the causal strength from $X$ to $Y$. For clearly, $\alpha_{1} =\lambda_{1}$ and $\alpha_{2} = \eta \lambda_{1} +
\lambda_{2}$ are the mixing coefficients of $L$ on $X$ and $Y$, respectively; $\beta_{1}$ and $\beta_{2}=\eta$ are the mixing coefficients of $E_X$ on $X$ and $Y$, respectively; and $\gamma_{2}$ is the mixing coefficients of $E_Y$ on $Y$. $E_X$ and $E_Y$ are two independent noise terms.

In the perspective of the matrix, Eq. (\ref{eq:figure2}) can be transformed into the form of an OICA model over two observed variables $X$ and $Y$ as:
\begin{equation}
\begin{aligned}
\underbrace{
\begin{bmatrix}
X\\
Y\\
\end{bmatrix}
}_{\mathbf{V}}
&=
\underbrace{
\begin{bmatrix}
    0 & 0 \\
    \eta & 0 \\
\end{bmatrix}
}_{\mathbf{B}}
\underbrace{
\begin{bmatrix}
X\\
Y\\
\end{bmatrix}
}_{\mathbf{V}}
+
\underbrace{
\begin{bmatrix}
    \lambda_1\\
    \lambda_2\\
\end{bmatrix}
}_{\mathbf{\Lambda}}
\underbrace{
\begin{bmatrix}
L\\
\end{bmatrix}
}_{\mathbf{L}}
+
\underbrace{
\begin{bmatrix}
E_X\\
E_Y\\
\end{bmatrix}
}_{\mathbf{E}} \\
&=
\underbrace{
\begin{bmatrix}
    \alpha_1 & \beta_1 & 0 \\
    \alpha_2 & \beta_2 & \gamma_2 \\
\end{bmatrix}
}_{\mathbf{A}}
\underbrace{
\begin{bmatrix}
L\\
E_X\\
E_Y\\
\end{bmatrix}
}_{\mathbf{S}}
.
\end{aligned}
\label{eq:mixing}
\end{equation}

\subsection{Cumulants}
The cumulants is a measure to capture the (joint) probability distribution information from data. The definition of cumulants of a random vector $X$ is formalized as:
\begin{definition}
[\textbf{Cumulants} \citep{brillinger2001time}] Let $X=(X_1, X_2, \dots, X_n)$ be a random vector of length $n$. The $k$-th order cumulant tensor of $X$ is defined as a $n \times \cdots  \times n$ ($k$ times) table, $\mathcal{C}^{(k)}$, whose entry at position ($i_1, \cdots, i_k$) is

\begin{equation}
    \begin{aligned}
    &\mathcal{C}^{(k)}_{i_1 \!, \cdots \!,  i_k}  = Cum(X_{i_1}, \dots, X_{i_k}) \\
    &=  \sum_{(D_1\!, \dots \!, D_h)} (-1)^{h \!- \!1}(h \!- \!1)! \mathbb{E}\!\left[ \! \prod_{j\in D_i}X_j  \! \right] \cdots \mathbb{E} \!\left[ \! \prod_{j \in D_h} \! X_j \! \right]\!,
    \end{aligned}
\end{equation}
where the sum is taken over all partitions $(D_1, \dots, D_h)$ of the set $\{i_1, \dots, i_k\}$. 
\end{definition}
For convenience, we use $\mathcal{C}_{i}(X)$ to denote $Cum(\underbrace{X, \dots}_{i \,\text{times}})$, and use  $\mathcal{C}_{i, j}(X, Y)$ to denote $Cum(\underbrace{X, \dots}_{i \,\text{times}}, \underbrace{Y, \dots}_{j \,\text{times}})$. For example, $\mathcal{C}_{5}(X)$ represents $Cum(X, X, X, X, X)$, $\mathcal{C}_{2, 3}(X, Y)$ represents $Cum(X, X, Y, Y, Y)$.

Note that the first-order cumulant is the mean, the second-order cumulant is the variance, and the third-order cumulant is the same as the third central moment. If each variable $X_i$ has zero mean, then the sum of the partitions with size 1 is 0 and can be omitted. For example, for $X$ and $Y$ that are generated by Eq. (\ref{eq:mixing}), the 3rd and 5th order cumulants or joint cumulants of $X$ and $Y$ are:
\begin{equation}
    \begin{aligned}
    \mathcal{C}_3(X) &= \alpha_1^3 \mathcal{C}_3(L) + \beta_1^3 \mathcal{C}_3(E_X), \\
    \mathcal{C}_{2,1}(X, Y) &= \alpha_1^2\alpha_2 \mathcal{C}_3(L) + \beta_1^2 \beta_2  \mathcal{C}_3(E_X), \\
    \mathcal{C}_{1,2}(X, Y) &= \alpha_1\alpha_2^2 \mathcal{C}_3(L) + \beta_1 \beta_2^2 \mathcal{C}_3(E_X), \\
    \mathcal{C}_3(Y) &= \alpha_2^3 \mathcal{C}_3(L) + \beta_2^3 \mathcal{C}_3(E_X) + \gamma_2^3 \mathcal{C}_3(E_Y),\\
    \mathcal{C}_{2,3}(X, Y) &= \alpha_1^2 \alpha_2^3 \mathcal{C}_3(L) + \beta_1^2 \beta_2^3 \mathcal{C}_3(E_X),\\
    \mathcal{C}_{3,2}(X, Y) &= \alpha_1^3 \alpha_2^2 \mathcal{C}_3(L) + \beta_1^3 \beta_2^2 \mathcal{C}_3(E_X).
    \end{aligned}
    \label{eq:figbcum}
\end{equation}

The equations above reveal that the higher-order cumulants imply the property of the generation process of the observed data (even when latent variables are present).

\section{Intuition}\label{sec:intuition}
The intuition of our method is based on the following observations. Take the causal graph between two observed variables $X$ and $Y$, given in Figure \ref{fig:example}(a) as an example. $X$ and $Y$ are affected by a latent variable, and there exists no causal directed edge between them. The joint cumulant of $X$ and $Y$ can be approximately regarded as the multiplication of powers of coefficients and the cumulants of their shared components, which can be expressed as follows:
\begin{equation}
    \begin{aligned}
    \mathcal{C}_{1,3}(X, Y) &= \lambda_1\lambda_2^3\mathcal{C}_{4}(L),\\
    \mathcal{C}_{2,2}(X, Y) &= \lambda_1^2\lambda_2^2\mathcal{C}_{4}(L),\\
    \mathcal{C}_{1,3}(X, Y) &= \lambda_1^3\lambda_2\mathcal{C}_{4}(L),\\
    \end{aligned}
\end{equation}
where the joint cumulants only have a term about $\mathcal{C}_{4}(L)$. If multiply the above joint cumulants, one can obtain
\begin{equation}
   \mathcal{C}_{2,2}(X,Y)^2 = \mathcal{C}_{1, 3}(X,Y)\mathcal{C}_{3,1}(X, Y).
   \label{eq:ext_tetrad_constraint_example}
\end{equation}
When there exists a causal directed edge between two observed variables, e.g., as in Figure \ref{fig:example}(b), the joint cumulant contains two different terms about $\mathcal{C}_{4}(L)$ and $\mathcal{C}_{4}(E_X)$ are:
\begin{equation}
    \begin{aligned}
    \mathcal{C}_{1,3}(X,Y) &= \lambda_1(\lambda_1\eta + \lambda_2)^{3}\mathcal{C}_{4}(L) + \eta^{3}\mathcal{C}_{4}(E_X),\\
    \mathcal{C}_{2,2}(X,Y) &= \lambda_1^2(\lambda_1\eta + \lambda_2)^{2}\mathcal{C}_{4}(L) + \eta^{2}\mathcal{C}_{4}(E_X),\\
    \mathcal{C}_{3,1}(X,Y) &= \lambda_1^3(\lambda_1\eta + \lambda_2)\mathcal{C}_{4}(L) + \eta^{1}\mathcal{C}_{4}(E_X).\\
    \end{aligned}
\end{equation}
If we compute the constraint as Eq. (\ref{eq:ext_tetrad_constraint_example}), the term $\lambda_1^2(\lambda_1\eta + \lambda_2)^{2}\eta^{2}\mathcal{C}_{4}(Y)\mathcal{C}_{4}(E_X)$ appears on the left-hand side of Eq. (\ref{eq:ext_tetrad_constraint_example}), but it is absent from the right-hand side. This discrepancy leads to the violation of Eq. (\ref{eq:ext_tetrad_constraint_example}).

Furthermore, if we find the violation of the Eq. (\ref{eq:ext_tetrad_constraint_example}), that means there exists a causal edge between two observed variables. The following question is how to determine its causal direction. Consider the causal graph given in Figure \ref{fig:example}(b), where $X$ is a parent of $Y$ and both of them are affected by a latent confounder $L$. The data generation process of $X$ and $Y$ is described by Eq. (\ref{eq:figure2}). Note that in this case, $X$ and $Y$ share two latent independent components, namely $L$ and $E_X$. Next, we will show that by utilizing fifth-order cumulants, the causal direction between two observed variables can be identified.

First, considering the covariance (i.e., the second-order cumulant), the second-order (joint) cumulants are:
\begin{equation}
    \begin{aligned}
    Var(X) &= \alpha_1^2 Var(L) + \beta_1^2 Var(E_X), \\
    Cov(X, Y) &= \alpha_1\alpha_2 Var(L) + \beta_1 \beta_2 Var(E_X), \\
    Var(Y) &= \alpha_2^2 Var(L) + \beta_2^2 Var(E_X) + \gamma_2^2 Var(E_Y).
    \end{aligned}
\end{equation}

We observed that the variance of the cause variable $X$ only contains all the variance of shared components, e.g., $Var(L)$ and $Var(E_X)$, but not the effect variable $Y$. It shows that the parameter estimation of the shared components sheds light on the identification of causal direction. Although we can find that $Var(E_Y)$ is only in $Var(Y)$, we cannot uniquely determine the variance of $L$, $E_X$ and $E_Y$, and estimate all the parameters by using only the three equations above, as the result in \cite{Ledermann1937OnTR, Bekker1997GenericGI}. 

Note that the covariance is the second-order cumulant, the idea that naturally arises is whether higher-order cumulants can be used to solve the problem. Thus, we consider the third order cumulant of $X$ and $Y$: 
\begin{equation}
    \begin{aligned}
    \mathcal{C}_{3}(X) &= \alpha_1^3 \mathcal{C}_{3}(L) + \beta_1^3 \mathcal{C}_{3}(E_X), \\
    \mathcal{C}_{3}(Y) &= \alpha_2^3 \mathcal{C}_{3}(L) + \beta_2^3 \mathcal{C}_{3}(E_X)+ \gamma_2^3 \mathcal{C}_{3}(E_Y).\\
    \end{aligned}
\end{equation}

Suppose we can estimate the terms $\alpha_1^3\mathcal{C}_{3}(L)$, $\beta_1^3  \mathcal{C}_{3}(E_X)$, $\alpha_2^3\mathcal{C}_{3}(L)$, and $\beta_2^3 \mathcal{C}_{3}(E_X)$, then we have the asymmetries that
\begin{equation}
\begin{aligned}
\mathcal{R}_{X \to Y} = \mathcal{C}_{3}(X) - \alpha_1^3\mathcal{C}_{3}(L) - \beta_1^3  \mathcal{C}_{3}(E_X) = 0, \\
\mathcal{R}_{Y\to X} = \mathcal{C}_{3}(Y) - \alpha_2^3\mathcal{C}_{3}(L) - \beta_2^3 \mathcal{C}_{3}(E_X) \neq 0.
\end{aligned}
\end{equation}

Interestingly, by using higher-order cumulants, the term $\alpha_1^3 \mathcal{C}_{3}(L)$ can be obtained by the following equation through the estimated $\frac{\alpha_{2}}{\alpha_{1}}$ and $\frac{\beta_2}{\beta_1}$:
\begin{equation}
\alpha_1^3 \mathcal{C}_{3}(L) = \frac{\mathcal{C}_{1,2}(X, Y) - \frac{\beta_2}{\beta_1} \times \mathcal{C}_{2,1}(X, Y)}{\frac{\alpha_{2}}{\alpha_{1}} \times (\frac{\alpha_{2}}{\alpha_{1}}-\frac{\beta_2}{\beta_1})} , 
\end{equation}
where $\frac{\alpha_{2}}{\alpha_{1}}\neq \frac{\beta_2}{\beta_1}$. The methods of estimating $\frac{\alpha_{2}}{\alpha_{1}}$ and $\frac{\beta_2}{\beta_1}$ will be introduced in the next subsection.
Similarly, $\beta_1^3 \mathcal{C}_{3}(E_X)$, $\alpha_2^3 \mathcal{C}_{3}(L)$, and $\beta_2^3 \mathcal{C}_{3}(E_X)$ can be obtained. Roughly speaking, we can determine the causal relationship only by utilizing the higher-order cumulants. Note that the insights into inferring causal relationships between two observed variables can be attained not only through third-order cumulants but also by harnessing cumulant orders higher than three.

\section{Proposed Method}
\subsection{Parameters Estimation}

To obtain $\mathcal{R}_{X \to Y}$ and $\mathcal{R}_{Y\to X}$, we need to first estimate the parameters on the ratio of mixing coefficients $\frac{\alpha_{2}}{\alpha_{1}}$ and $\frac{\beta_2}{\beta_1}$. 

Given the causal structure in Figure \ref{fig:example}(b), we can obtain the fifth-order (joint) cumulants of $X$ and $Y$ as:
\begin{equation}
    \begin{aligned}
    \mathcal{C}_{4, 1}(X, Y) &= \alpha_{1}^{4}\alpha_{2} \mathcal{C}_{5}(L) + \beta_{1}^{4}\beta_{2} \mathcal{C}_{5}(E_X),\\
    \mathcal{C}_{3, 2}(X, Y) &= \alpha_{1}^{3}\alpha_{2}^{2} \mathcal{C}_{5}(L) + \beta_{1}^{3}\beta_{2}^{2} \mathcal{C}_{5}(E_X),\\
    \mathcal{C}_{2, 3}(X, Y) &= \alpha_{1}^{2}\alpha_{2}^{3} \mathcal{C}_{5}(L) + \beta_{1}^{2}\beta_{2}^{3} \mathcal{C}_{5}(E_X),\\
    \mathcal{C}_{1, 4}(X, Y) &= \alpha_{1}\alpha_{2}^{4} \mathcal{C}_{5}(L) + \beta_{1}\beta_{2}^{4} \mathcal{C}_{5}(E_X).\\
    \end{aligned}
\end{equation}

Note that in the above equations, $\alpha_1$, $\alpha_2$, $\beta_1$ and $\beta_2$ are not zero. Let $\theta_\alpha = \frac{\alpha_2}{\alpha_1}$ and $\theta_\beta=\frac{\beta_2}{\beta_1}$. Then, by combining the above equations, we can obtain $\theta_\alpha$ and $\theta_\beta$ by an analytical solution of the following quadratic equation directly:
\begin{equation}
    \begin{aligned}
    & \left( \mathcal{C}_{3,2}(X, Y)^2 - \mathcal{C}_{2, 3}(X, Y) \mathcal{C}_{4, 1}(X, Y) \right) \theta^{2} \\
    + & \left( \mathcal{C}_{1,4}(X, Y)\mathcal{C}_{4,1}(X, Y) - \mathcal{C}_{2, 3}(X, Y) \mathcal{C}_{3, 2}(X, Y) \right)\theta\\
    + & \mathcal{C}_{2,3}(X, Y)^2 - \mathcal{C}_{1, 4}(X, Y) \mathcal{C}_{3, 2}(X, Y) \\
    = & 0.    
    \end{aligned}
    \label{eq:a_2}
\end{equation}
The solution to the above equation is as follows:
\begin{equation}
    a\theta^2 + b\theta + c = 0 \Rightarrow \theta^{*} = \frac{-b\pm\sqrt{b^2-4ac}}{2a}
    \label{eq:analytics},
\end{equation}
where $a=\mathcal{C}_{3,2}(X, Y)^2 - \mathcal{C}_{2, 3}(X, Y) \mathcal{C}_{4, 1}(X, Y) \neq 0$, $b=\mathcal{C}_{1,4}(X, Y)\mathcal{C}_{4,1}(X, Y) - \mathcal{C}_{2, 3}(X, Y) \mathcal{C}_{3, 2}(X, Y)$, and $c=\mathcal{C}_{2,3}(X, Y)^2 - \mathcal{C}_{1, 4}(X, Y) \mathcal{C}_{3, 2}(X, Y)$. Practically, we cannot obtain the exact value of $\theta_\alpha$ and $\theta_\beta$, but they are one of the element in ${\frac{-b + \sqrt{b^2-4ac}}{2a}, \frac{-b - \sqrt{b^2-4ac}}{2a}}$ and $\theta_\alpha \neq \theta_\beta$. This result can still help us infer causal relationships.

The process of deriving a quadratic equation is provided in the appendix. Note that other cumulants higher than order 5 can also be used to estimate parameters. Thus, we can conclude the theorem of parameters estimation by using generalized order cumulants as follows: 

\begin{theorem} Assume that two observed variables $X$ and $Y$ are generated by Eq. (\ref{eq:model}). Suppose there exists $i, j > 0$, such that $\mathcal{C}_{i+3, j}(X, Y)$, $\mathcal{C}_{i+2, j+1}(X, Y)$, $\mathcal{C}_{i+1, j+2}(X, Y)$ and $\mathcal{C}_{i, j+3}(X, Y)$ are not equal to zero. If $X$ and $Y$ are affected by two independent components with different ratios of mixing coefficients, then the ratio of mixing coefficients $\theta_1$ and $\theta_2$ of two independent components on $X$ and $Y$ can be estimated by the analytical solution of the following equation:
\begin{equation}
\begin{aligned}
    &\left( \mathcal{C}_{i+2, j+1}(X, Y)^2 - \mathcal{C}_{i+1, j+2}(X, Y) \mathcal{C}_{i+3, j}(X, Y) \right) \theta^{2} \\
    + & \mathcal{C}_{i, j+3}(X, Y)\mathcal{C}_{i+3, j}(X, Y) \theta \\
    -& \mathcal{C}_{i+1, i+2}(X, Y) \mathcal{C}_{i+2, j+1}(X, Y) \theta\\
    + & \mathcal{C}_{i+1, j+2}(X, Y)^2 - \mathcal{C}_{i, j+3}(X, Y) \mathcal{C}_{i+2, j+1}(X, Y) \\
    = & 0.
\end{aligned}
\label{eq:general_quadratic_equation}
\end{equation}
\end{theorem}

Note that $\theta_1$ and $\theta_2$ can be obtained using cumulant orders higher than five. However, this study demonstrates the minimum required higher-order cumulant for parameter estimation, highlighting that at least fifth-order cumulants are necessary for accurate parameter estimation. 

Because Eq. (\ref{eq:general_quadratic_equation}) is a quadratic equation, its solution would fail when $\mathcal{C}_{i+2, j+1}(X, Y)^2 - \mathcal{C}_{i+1, j+2}(X, Y) \mathcal{C}_{i+3, j}(X, Y) = 0$. Interestingly, when this condition holds, we find that these two observed variables are affected by the same latent variable, and they are not the cause of each other. This condition also can be seen as a specific case in Eq. (\ref{eq:no_edge}), which can be proved in the following section. Besides, when there is no directed edge between $X$ and $Y$ that are influenced by one latent variable, we can estimate the mixing coefficients of the latent variable on $X$ and $Y$, denoted as $\hat{\alpha}_1$ and $\hat{\alpha}_2$, by:
\begin{equation}
    \begin{aligned}
        \hat{\alpha}_1 &= \sqrt{\frac{\mathcal{C}_{i+1, j}(X, Y)}{\mathcal{C}_{i, j+1}(X, Y)} \cdot \mathcal{C}_{1, 1}(X, Y)}, \\
        \hat{\alpha}_2 &= \frac{\mathcal{C}_{1, 1}(X, Y)}{\hat{\alpha}_1},
    \end{aligned}
    \label{eq:shared-one-exogenous}
\end{equation}
which have been proved in the work \cite{cai2023causal}.

\subsection{Identifiability}
Based on the above analysis, we provide the identification results for the causal structure over two observed variables with one latent variable in this section.

Considering two observed variables that are affected by one latent confounder, we can detect whether there is a directed edge between them by the following theorem.

\begin{theorem}
Assume that two observed variables $X$ and $Y$ are generated by Eq. (\ref{eq:model}), and $X$ and $Y$ are affected by the same latent variable $L$. Suppose there exists $i, j > 0$, such that $\mathcal{C}_{i, j+2}(X, Y)$, $\mathcal{C}_{i+1, j+1}(X, Y)$ and $\mathcal{C}_{i+2, j}(X, Y)$ are not equal to zero. Then there is no directed edge between $X$ and $Y$, if and only if $\mathcal{C}_{i+1, j+1}(X, Y)^2 - \mathcal{C}_{i, j+2}(X, Y) \mathcal{C}_{i+2, j}(X, Y) = 0$.  
\label{th:edge_existence}
\end{theorem}

Theorem \ref{th:edge_existence} provides a method for identifying whether there exists a directed edge between two observed variables in the presence of latent variables. If such a directed edge between two variables exists, a further question is how to identify the direction of the edge.

Fortunately, from the intuition provided in Section \ref{sec:intuition}, we can solve this problem. Because we have no idea of the causal direction, we introduce $S_1$ and $S_2$ to be two shared independent components of $X$ and $Y$. Let $\mathcal{R}_{X \to Y}=\mathcal{C}_{3}(X) - \alpha_1^3 \mathcal{C}_{3}(S_1) - \beta_1^3  \mathcal{C}_3(S_2)$ (and $\mathcal{R}_{Y \to X}=\mathcal{C}_{3}(Y) - \alpha_2^3 \mathcal{C}_{3}(S_1) - \beta_2^3  \mathcal{C}_3(S_2)$) be the causal direction criteria. Then we can obtain the following rule: if and only if $\mathcal{R}_{X \to Y}=0$, then $X$ is a parent of $Y$ and both of them are affected by latent confounder. This rule can be summarized as:

\begin{theorem}
Assume that two observed variables $X$ and $Y$ are generated by Eq. (\ref{eq:model}), and $X$ and $Y$ are affected by the same latent variable $L$. Then $X$ is a cause of $Y$ if and only if $\mathcal{R}_{X \to Y} = 0$. 
\label{th:causal_direction}
\end{theorem}

Theorem \ref{th:causal_direction} provides a method for causal direction identification, achieved by the higher-order cumulants. Combining Theorem \ref{th:edge_existence} and Theorem \ref{th:causal_direction}, we can identify the causal relationship between two observed variables that are affected by a latent confounder, which is guaranteed by the following theorem.
\begin{theorem}
Assume that two observed variables $X$ and $Y$ are generated by Eq. (\ref{eq:model}), and $X$ and $Y$ are affected by the same latent variable $L$. The causal relationship between two observed variables can be identified by using the higher-order cumulants. 
\label{th:identifiable_model}
\end{theorem}

The principles outlined in Theorem \ref{th:identifiable_model} provide a foundation for identifying causal structures within a two-variable context. It can be extended to scenarios involving more than two variables by transforming the causal structure into a canonical model, where each latent variable is the parent of two observed variables. Consequently, the criterion for identifying causal structures encompassing multiple variables stipulates that any pair of observed variables can be influenced by a maximum of one latent variable. 

\subsection{Learning Algorithm}
Based on the identifiability results, we now consider a practical method for inferring causal structure from observed data with latent confounders.

We begin by considering the case where two observed variables $X$ and $Y$ with one latent confounder. Denote the statistic $s = \mathcal{C}_{3,2}(X, Y)^2 - \mathcal{C}_{2, 3}(X, Y) \mathcal{C}_{4, 1}(X, Y)$ as Theorem \ref{th:edge_existence}. It is straightforward that, first, test whether $s$ is equal to zero to identify the presence of a directed edge between them. If such an edge exists, we further determine its direction by using Theorem \ref{th:causal_direction}. 

In practice, the statistic wouldn't equal zero exactly, since the sample size is finite. To test whether the statistic $s$ is equal to zero, we provide a procedure as following steps: 

1) Sample data of size $m$ from the original dataset $(\mathbf{X}, \mathbf{Y})$ with replacement.

2) Compute the statistic $s$. 

3) Repeat steps 1 and 2 for $B$ times to obtian $\lbrace s_{m,1}^{*}, \dots, s_{m,B}^{*} \rbrace$.

4) Perform one sample sign test \cite{dixon1946statistical} on $\lbrace s_{m,1}^{*}, \dots, s_{m,B}^{*} \rbrace$.

Similarly, we can use the above procedure to test whether $R_{X \to Y}$ is equal to zero. The reason we do this is that Non-Gaussian distribution contains a wide class of distribution, we can not find suitable statistics to capture the information of distribution without any prior information of distribution. Further, we conjecture the median can be asymptotically approximated to the mean, so we use the sign test to test whether the median of $\lbrace s_{m,1}^{*}, \dots, s_{m,B}^{*} \rbrace$ is equal to zero.

The above procedure will result in one of several possible scenarios. First, if $X$ and $Y$ are only affected by a latent confounder, then we can infer that there is no directed edge between the two, and no further analysis is performed. Second, if there exists a directed edge between them, but both directional edges are accepted. We conclude that either model may be correct but we cannot infer it from the data. The positive result is when we are able to reject one of the directions and accept the other.

\section{Simulations}

\subsection{Experiment Setups}
To evaluate the performance of our proposed method, we conducted experiments on synthetic data, considering the following two cases: 

[Case 1]: A causal graph over two observed variables that are affected by a latent variable, and there is no causal directed edge between observed variables, i.e., Figure \ref{fig:example}(a). 

[Case 2]: A causal graph over two observed variables that are affected by a latent variable, and there is a causal directed edge between observed variables, i.e., Figure \ref{fig:example}(b).

The purposes of these experiments are: (1) Figure \ref{fig:example}(a) is used to evaluate the performance on inferring the existence of the causal directed edge of different methods; (2) Figure \ref{fig:example}(b) is used to evaluate the performance on inferring the direction of the causal directed edge of different methods.

Given the causal graphs, we randomly generated data according to Eq. (\ref{eq:model}), where the causal coefficient is sampled from a uniform distribution between $[0.8, 1.2]$. The noise terms are generated from three distinct distributions: Exponential Distribution, Gamma Distribution with shape parameter $k=3$ and Gumbel distribution. For each model, the sample size $N$ is varied among $\lbrace 5000, 10000, 50000, 100000 \rbrace$. For each setting, we generate 100 datasets.

\begin{figure*}[t]
\subfigure[Exponential Distribution]{
\includegraphics[width=0.32\textwidth]{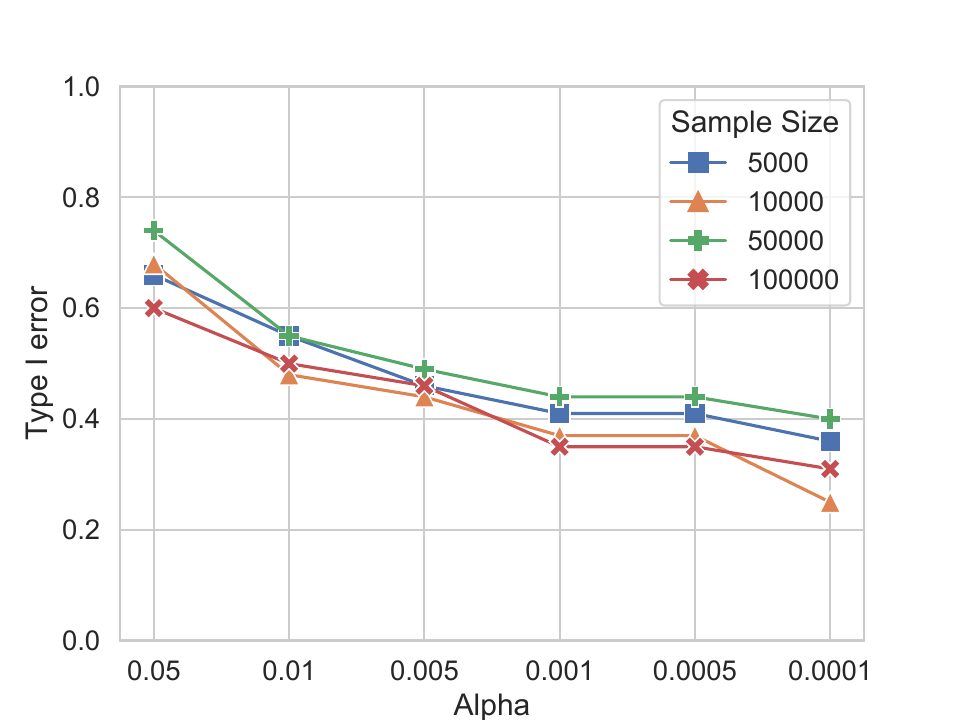}
}
\subfigure[Gamma Distribution]{
\includegraphics[width=0.32\textwidth]{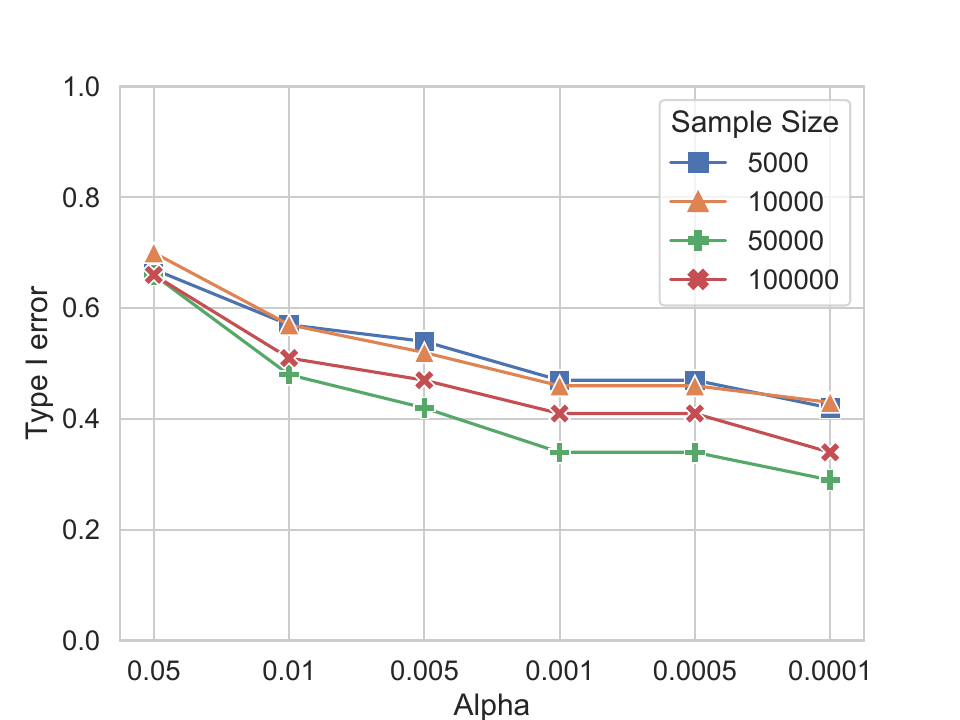}
}
\subfigure[Gumbel Distribution]{
\includegraphics[width=0.32\textwidth]{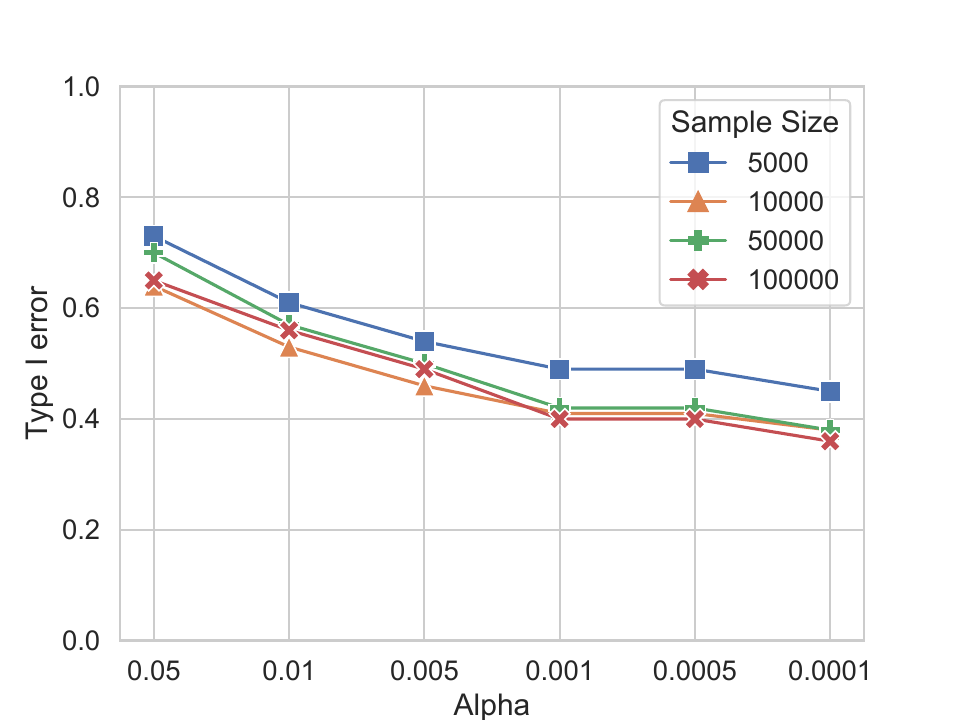}
}
\centering
\caption{The probability of Type I error varies across different distributions in the test for the presence of a causal edge.}
\label{fig:type1error}
\end{figure*}

\begin{figure*}[t]
\subfigure[Exponential Distribution]{
\includegraphics[width=0.32\textwidth]{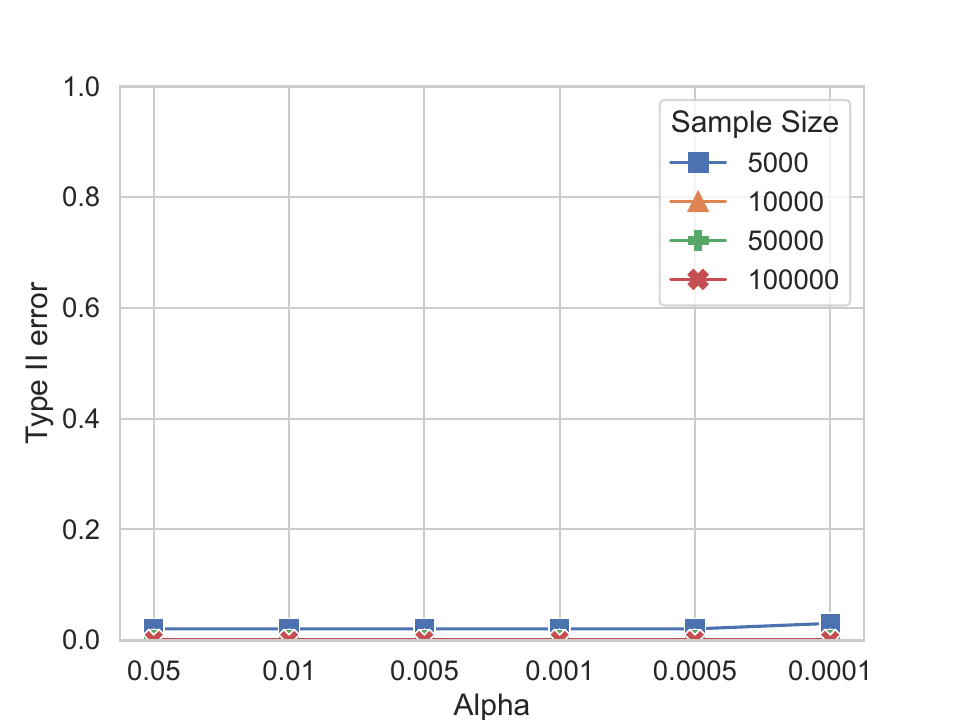}
}
\subfigure[Gamma Distribution]{
\includegraphics[width=0.32\textwidth]{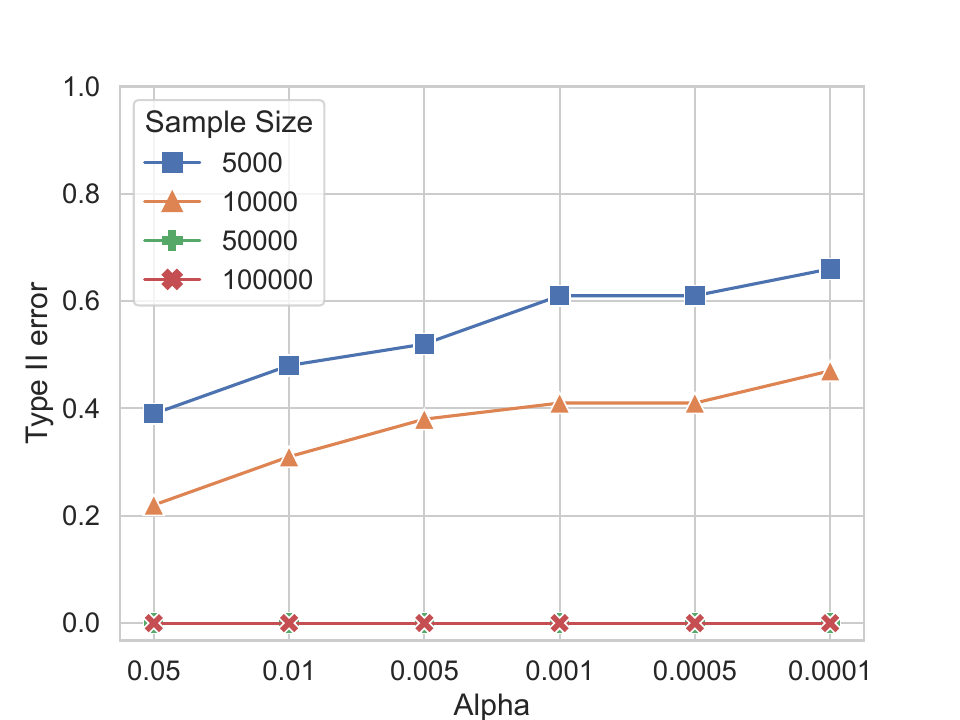}
}
\subfigure[Gumbel Distribution]{
\includegraphics[width=0.32\textwidth]{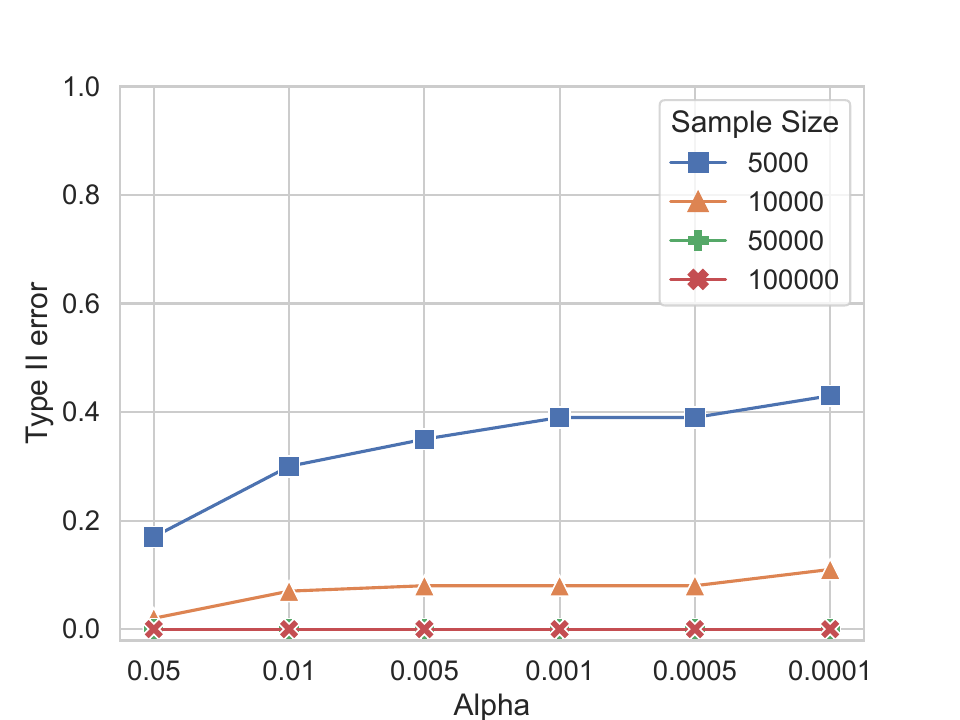}
}
\centering
\caption{The probability of Type II error varies across different distributions in the test for the presence of a causal edge.}
\label{fig:type2error}
\end{figure*}

\subsection{Baseline Methods and Evaluation Metrics}
In these experiments, we use Linear Non-Gaussian Acyclic Model (LiNGAM) \cite{shimizu2006linear}, Additive Noise Model (ANM) \cite{hoyer2008nonlinear}, LvLiNGAM \cite{hoyer2008estimation} and pairwise LvLiNGAM \cite{entner2011discovering} as the baseline methods. Among these methods, LiNGAM and ANM are two typical methods under the causal sufficiency assumption. LiNGAM leverages the ICA technique to estimate the mixing matrix and transformed it into the causal strength matrix, while ANM leverages the independence between the regression residual and the assumed cause to determine the causal direction. LvLiNGAM and Pairwise LvLiNGAM are two methods against the latent variables. Pairwise LvLiNGAM aims to capture partial information regarding causal relationships between variable pairs within data generated by LvLiNGAM. However, pairwise LvLiNGAM does not yield informative results when applied to synthetic data. Consequently, we only provide the results of LiNGAM, ANM, LvLiNGAM, and our proposed methods. For the test of our approach, we set $m = 0.8n$ and $B = 30$, where $n$ is the sample size of the initial dataset.

To evaluate the performance of different methods, we employ the accuracy score that the learned graph is the same as the true one as an evaluation metric. Besides, we also examine the probabilities of Type I error and Type II error of our hypothesis test procedure within specific simulation scenarios as sample size and significance level ($\alpha$ =$0.05$, $0.01$, $0.005$, $0.001$, $0.0005$ and $0.0001$) change.

\subsection{Experimental Results}

\begin{figure*}[t]
\subfigure[Exponential Distribution]{
\includegraphics[width=0.32\textwidth]{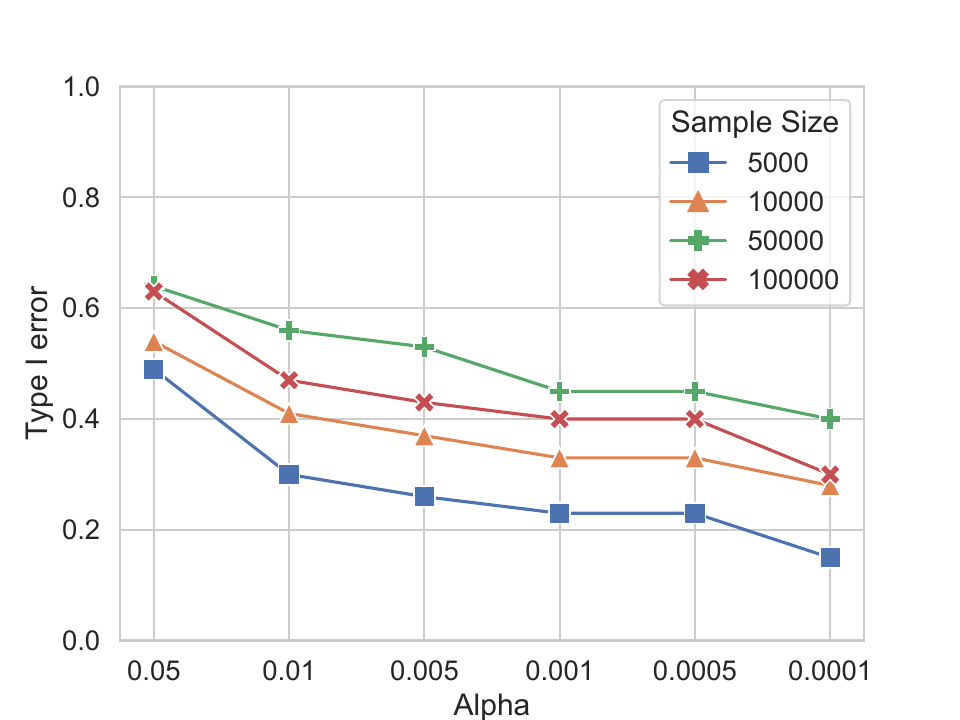}
}
\subfigure[Gamma Distribution]{
\includegraphics[width=0.32\textwidth]{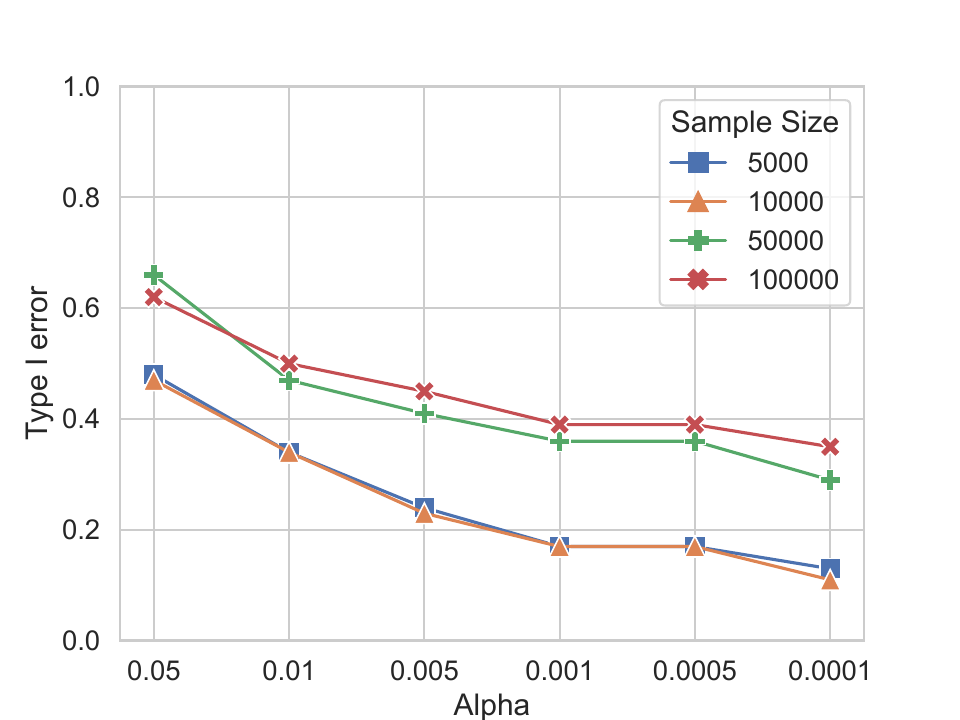}
}
\subfigure[Gumbel Distribution]{
\includegraphics[width=0.32\textwidth]{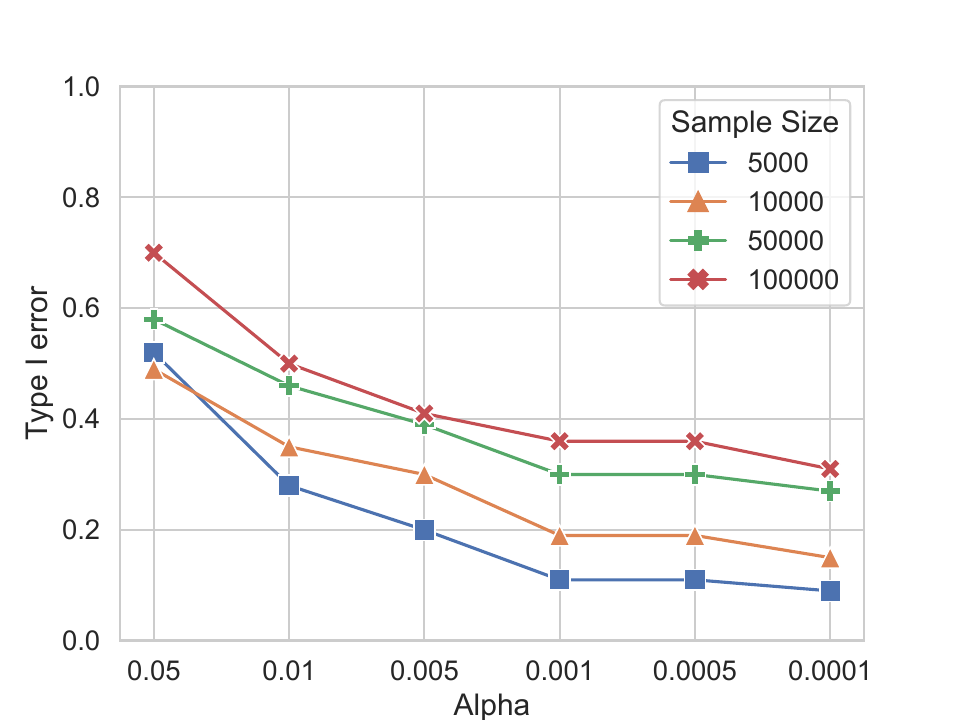}
}
\centering
\caption{The probability of Type I error varies across different distributions in the test for the direction of the causal edge.}
\label{fig:direction_type1error}
\end{figure*}

\begin{figure*}[t]
\subfigure[Exponential Distribution]{
\includegraphics[width=0.32\textwidth]{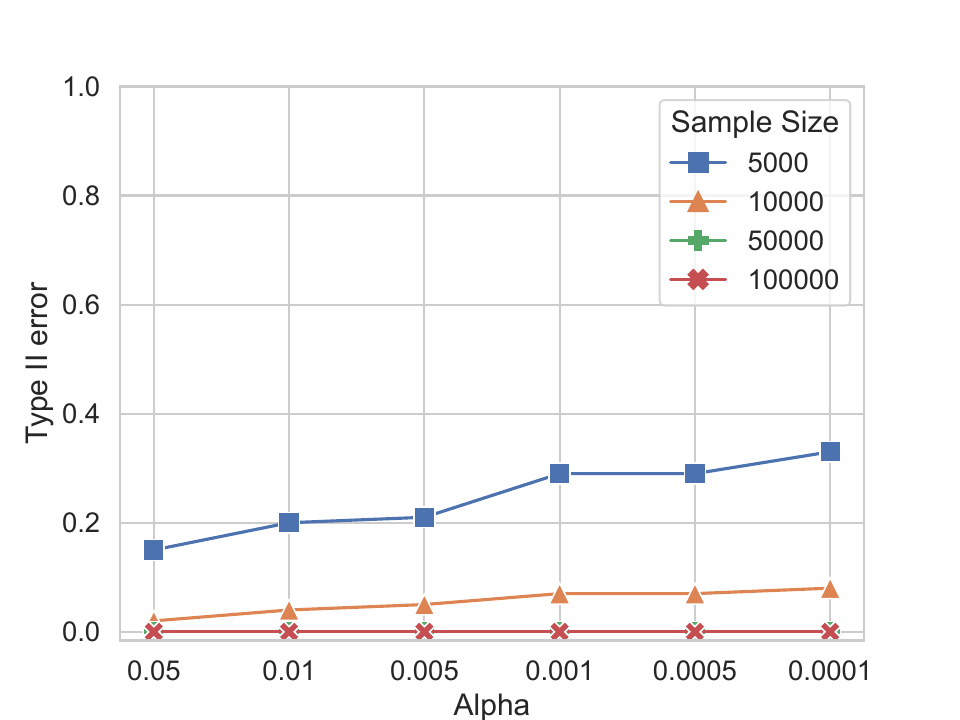}
}
\subfigure[Gamma Distribution]{
\includegraphics[width=0.32\textwidth]{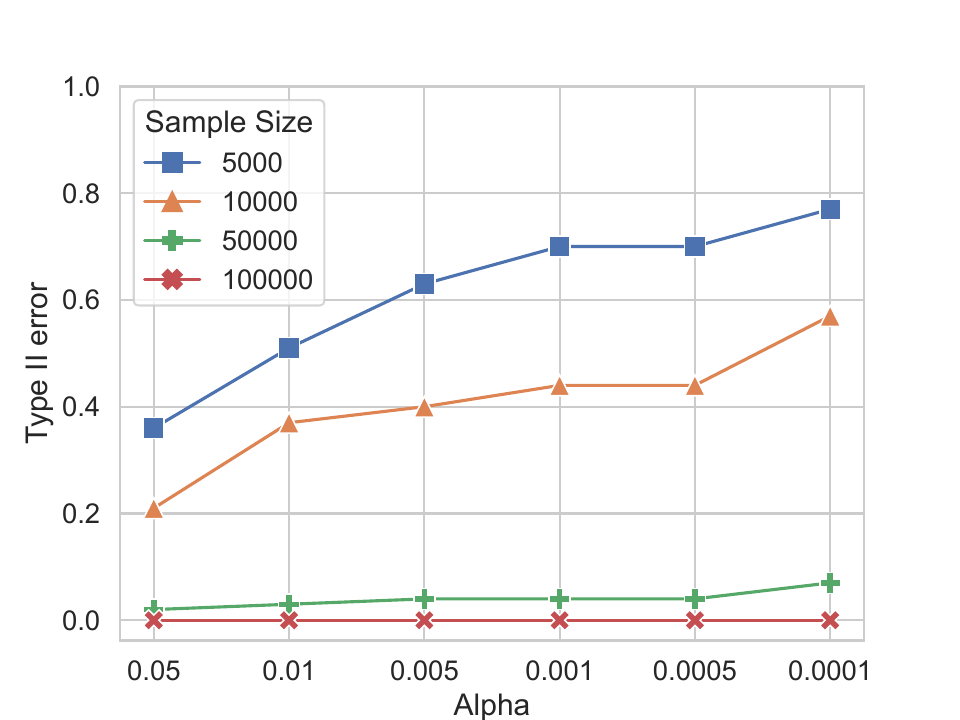}
}
\subfigure[Gumbel Distribution]{
\includegraphics[width=0.32\textwidth]{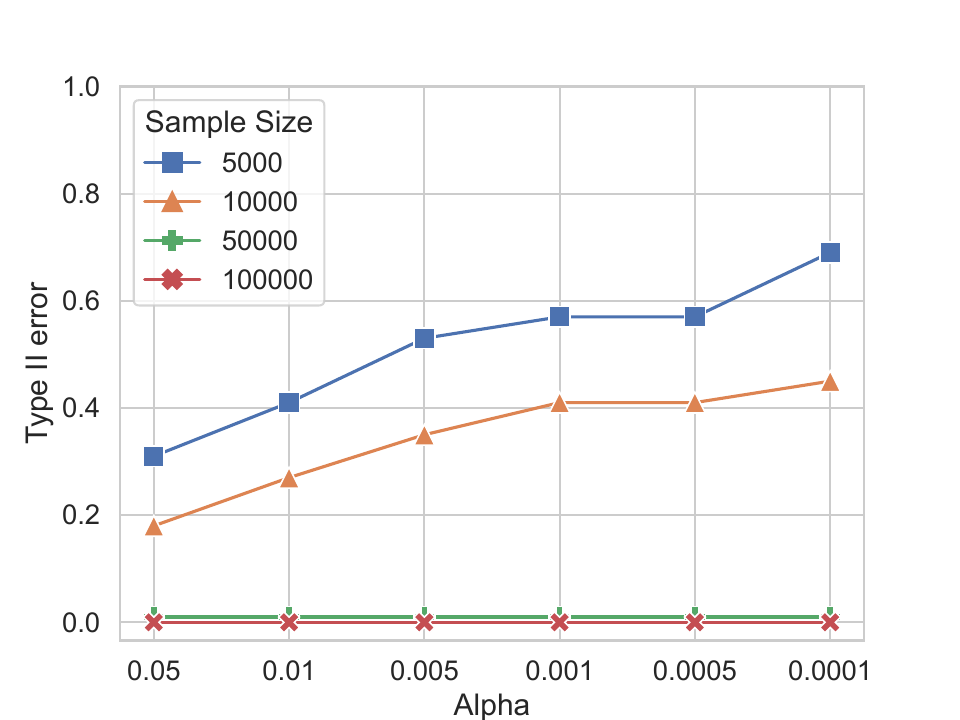}
}
\centering
\caption{The probability of Type II error varies across different distributions in the test of the direction of the causal edge.}
\label{fig:direction_type2error}
\end{figure*}

\paragraph{Evaluation in Case 1.}
The experimental results of case 1 are given in Table \ref{tab:summary-no-edge}. In this case, we just compare the result between LvLiNGAM and our methods due to ANM can not infer the existence of the causal directed edge. From the results, LvLiNGAM is unable to return the truth that there is no causal edge between observed variables. Because LvLiNGAM's performance depends on Overcomplete Independent Components Analysis, which usually gets stuck in local optima, it would infer redundant causal edges in practice. Our proposed method can determine that there is no edge between two observed variables in most cases, although the accuracy is not so high. These results rely on the sample size and the test method in practice, which are shown in the results on the Type I error and Type II errors.

\begin{table}[ht]
  \centering
    \resizebox{.99\linewidth}{!}{
    \begin{tabular}{cccccc}
    \toprule
          & Sample Size & LiNGAM & ANM   & LvLiNGAM & Ours \\
    \midrule
    \multirow{4}[2]{*}{Exp} & 5000  & 0.00  & -     & 0.00  & \textbf{0.64 } \\
          & 10000 & 0.00  & -     & 0.00  & \textbf{0.75 } \\
          & 50000 & 0.00  & -     & 0.00  & \textbf{0.60 } \\
          & 100000 & 0.00  & -     & 0.00  & \textbf{0.69 } \\
    \midrule
    \multirow{4}[2]{*}{Gamma} & 5000  & 0.00  & -     & 0.00  & \textbf{0.58 } \\
          & 10000 & 0.00  & -     & 0.00  & \textbf{0.57 } \\
          & 50000 & 0.00  & -     & 0.00  & \textbf{0.71 } \\
          & 100000 & 0.00  & -     & 0.00  & \textbf{0.66 } \\
    \midrule
    \multirow{4}[2]{*}{Gumbel} & 5000  & 0.00  & -     & 0.00  & \textbf{0.55 } \\
          & 10000 & 0.00  & -     & 0.00  & \textbf{0.62 } \\
          & 50000 & 0.00  & -     & 0.00  & \textbf{0.62 } \\
          & 100000 & 0.00  & -     & 0.00  & \textbf{0.64 } \\
    \bottomrule
    \end{tabular}
    }
    \caption{Accuracy of the different methods varies across different distributions in Case 1. Here ``-" denotes that the corresponding methods are not applicable.}
  \label{tab:summary-no-edge}
\end{table}

To examine the Type I error, we generated $X$ and $Y$ according to Figure \ref{fig:example}(a) in which there exists no causal directed edge between $X$ and $Y$. Figure \ref{fig:type1error} and Figure \ref{fig:type2error} show the resulting probability of Type I errors and that of Type II errors at different significance levels, respectively. From Figure \ref{fig:type1error}, we find that the Type I error is around $0.3$ under the significance level $\alpha = 0.0001$. The Type II error of our proposed method is sensitive to sample size. In detail, the Type II error is almost equal to zero even under the significance level $\alpha = 0.0001$ when the sample size is larger than 50000. So we use the results with a significance level of 0.0001 to compare baseline methods.

\begin{table}[ht]
  \centering
  \resizebox{.99\linewidth}{!}{
    \begin{tabular}{cccccc}
    \toprule
          & Sample Size & LiNGAM & ANM   & LvLiNGAM & Ours \\
    \midrule
    \multirow{4}[2]{*}{Exp} & 5000  & 0.05  & 0.47  & 0.50  & \textbf{0.79 } \\
          & 10000 & 0.05  & 0.13  & 0.66  & \textbf{0.88 } \\
          & 50000 & 0.06  & -     & 0.61  & \textbf{0.91 } \\
          & 100000 & 0.07  & -     & 0.65  & \textbf{0.92 } \\
    \midrule
    \multirow{4}[2]{*}{Gamma} & 5000  & 0.07  & 0.38  & 0.43  & \textbf{0.55 } \\
          & 10000 & 0.04  & 0.51  & 0.58  & \textbf{0.71 } \\
          & 50000 & 0.03  & -     & 0.52  & \textbf{0.85 } \\
          & 100000 & 0.02  & -     & 0.61  & \textbf{0.89 } \\
    \midrule
    \multirow{4}[2]{*}{Gumbel} & 5000  & 0.13  & 0.51  & 0.53  & \textbf{0.70 } \\
          & 10000 & 0.10  & 0.43  & 0.59  & \textbf{0.74 } \\
          & 50000 & 0.13  & -     & 0.51  & \textbf{0.90 } \\
          & 100000 & 0.05  & -     & 0.61  & \textbf{0.92 } \\
    \bottomrule
    \end{tabular}%
    }
    \caption{Accuracy of the different methods varies across different distributions in Case 2. Here ``-" denotes that the corresponding methods take too long to produce any results.}
  \label{tab:case2}
\end{table}

\paragraph{Evaluation in Case 2.} The results of Case 2 are shown in Table \ref{tab:case2}. For large sample sizes, HSIC is not applicable because of its high time consumption and memory consumption, so ANM cannot return any result when the sample sizes are 50000 and 100000. For fairness, we compare the p-value of the hypothesis test in both directions for ANM and our method, which direction of p-value is lower is the correct one. From the results, we can see that ANM and LvLiNGAM obtain an accuracy score of around 0.5 in all sample sizes. Because ANM doesn't take latent variables into account, which leads it cannot distinguish the causal direction. In the case of LvLiNGAM, its efficacy is intertwined with the Overcomplete ICA technique. However, this approach often encounters challenges by becoming trapped in local optima, leading to inaccurate results.

Besides, Figure \ref{fig:direction_type1error} - \ref{fig:direction_type2error} show the resulting probability of Type I errors and that of Type II errors at different significance levels, respectively. From the results, we can see that the Type I error increases when the sample size increase. The reason is that the hypothesis test tends not to reject the null hypothesis, which also leads to a high Type II error result. 

Furthermore, we applied our proposed method to the datasets that are generated by nonlinear causal relationships, to relax  the linear assumption. When considering an exponential noise distribution and a sample size of 100,000, our method achieved an accuracy of $0.84$ in determining the causal direction between two observed variables. This demonstrates the robustness and potential applicability of our approach in scenarios beyond linear relationships.

\section{Discussion and Conclusion}
In this paper, we provide the identifiability theories for inferring causal relationships between two observed variables with latent variables, by utilizing higher-order cumulants. Based on these identifiability theories, we derive a causal discovery method that first detects whether there exists an edge between two observed variables, and then determines the direction of the causal edge if such a causal edge exists. 

Compared with existing methods, the power of the identifiability results provided in this paper depends on the information from higher-order cumulants of non-Gaussian data. Interestingly, one can find the relations between our proposed criterion given in Eq. (\ref{eq:no_edge}) is similar with tetrad constraint \cite{silva2006learning} under the non-Gaussian assumption. The tetrad constraint is satisfied among four pure observed variables that are affected by one latent variable. Note that in Eq. (\ref{eq:no_edge}), when $i=1$ and $j=1$, one have $Cum(X, X, Y, Y)^2 = Cum(X, X, X, Y) Cum(X, Y, Y, Y)$. The combination of joint cumulants of $X$ and $Y$ can be used to eliminate the influence of the shared latent variable on them. To some extent, the satisfaction of this condition is facilitated by the non-Gaussian noise assumption and higher-order cumulants.

The experimental results reveal that our proposed method achieves favorable performance, particularly with larger sample sizes (around 100,000). This also reflects that the test method requires a large sample size to approximate the true joint cumulants of the variables. Notably, if we are able to find a distribution to approximate the combination of joint cumulants during the test process, it is plausible to design a more dependable test method that does not necessitate stringent sample size requirements. Moreover, one might consider the linear assumption to be overly restrictive and unsuitable for real-world scenarios. However, the experimental results obtained from data generated by additive nonlinear relationships demonstrate the potential applicability of our method even in nonlinear cases. If a more effective test can be devised, even with limited sample size, it would facilitate the extension of our method to high-dimensional scenarios in practical applications. This will be the research direction of our next work. 

\section*{Acknowledgments}

We sincerely appreciate the insightful discussions from Shenlong Pan and anonymous reviewers, which greatly helped to improve the paper. This research was supported in part by National Key R\&D Program of China (2021ZD0111501), Natural Science Foundation of China (62206064, 61876043, 61976052, 62206061), National Science Fund for Excellent Young Scholars (62122022), the major key project of PCL (PCL2021A12), Guangzhou Basic and Applied Basic Research Foundation (2023A04J1700). KZ was supported in part by the NSF-Convergence Accelerator Track-D award \#2134901, by the National Institutes of Health (NIH) under Contract R01HL159805, by grants from Apple Inc., KDDI Research, Quris AI, and IBT, and by generous gifts from Amazon, Microsoft Research, and Salesforce.

\bibliography{aaai24}

\begin{thebibliography}{26}
\providecommand{\natexlab}[1]{#1}

\bibitem[{Bekker and ten Josephus~Berge(1997)}]{Bekker1997GenericGI}
Bekker, P.~A.; and ten Josephus~Berge. 1997.
\newblock Generic global indentification in factor analysis.
\newblock \emph{Linear Algebra and its Applications}, 264: 255--263.

\bibitem[{Brillinger(2001)}]{brillinger2001time}
Brillinger, D.~R. 2001.
\newblock \emph{Time series: data analysis and theory}.
\newblock Society for Industrial and Applied Mathematics.

\bibitem[{Cai et~al.(2023)Cai, Huang, Chen, Hao, and Zhang}]{cai2023causal}
Cai, R.; Huang, Z.; Chen, W.; Hao, Z.; and Zhang, K. 2023.
\newblock Causal Discovery with Latent Confounders Based on Higher-Order Cumulants.
\newblock \emph{International conference on machine learning}.

\bibitem[{Chen et~al.(2021)Chen, Cai, Zhang, and Hao}]{chen2021causal}
Chen, W.; Cai, R.; Zhang, K.; and Hao, Z. 2021.
\newblock Causal Discovery in Linear Non-Gaussian Acyclic Model With Multiple Latent Confounders.
\newblock \emph{IEEE Transactions on Neural Networks and Learning Systems}.

\bibitem[{Colombo et~al.(2012)Colombo, Maathuis, Kalisch, and Richardson}]{colombo2014learning}
Colombo, D.; Maathuis, M.~H.; Kalisch, M.; and Richardson, T.~S. 2012.
\newblock {Learning high-dimensional directed acyclic graphs with latent and selection variables}.
\newblock \emph{The Annals of Statistics}, 40(1): 294 -- 321.

\bibitem[{Dixon and Mood(1946)}]{dixon1946statistical}
Dixon, W.~J.; and Mood, A.~M. 1946.
\newblock The statistical sign test.
\newblock \emph{Journal of the American Statistical Association}, 41(236): 557--566.

\bibitem[{Entner and Hoyer(2011)}]{entner2011discovering}
Entner, D.; and Hoyer, P.~O. 2011.
\newblock Discovering unconfounded causal relationships using linear non-gaussian models.
\newblock In \emph{New Frontiers in Artificial Intelligence: JSAI-isAI 2010 Workshops, LENLS, JURISIN, AMBN, ISS, Tokyo, Japan, November 18-19, 2010, Revised Selected Papers 2}, 181--195. Springer.

\bibitem[{Eriksson and Koivunen(2004)}]{eriksson2004identifiability}
Eriksson, J.; and Koivunen, V. 2004.
\newblock Identifiability, separability, and uniqueness of linear ICA models.
\newblock \emph{IEEE Signal Processing Letters}, 11(7): 601--604.

\bibitem[{Hoyer et~al.(2008{\natexlab{a}})Hoyer, Janzing, Mooij, Peters, and Sch{\"o}lkopf}]{hoyer2008nonlinear}
Hoyer, P.; Janzing, D.; Mooij, J.~M.; Peters, J.; and Sch{\"o}lkopf, B. 2008{\natexlab{a}}.
\newblock Nonlinear causal discovery with additive noise models.
\newblock \emph{Advances in neural information processing systems}, 21.

\bibitem[{Hoyer et~al.(2008{\natexlab{b}})Hoyer, Shimizu, Kerminen, and Palviainen}]{hoyer2008estimation}
Hoyer, P.~O.; Shimizu, S.; Kerminen, A.~J.; and Palviainen, M. 2008{\natexlab{b}}.
\newblock Estimation of causal effects using linear non-Gaussian causal models with hidden variables.
\newblock \emph{International Journal of Approximate Reasoning}, 49(2): 362--378.

\bibitem[{Huang et~al.(2022)Huang, Low, Xie, Glymour, and Zhang}]{huang2022latent}
Huang, B.; Low, C. J.~H.; Xie, F.; Glymour, C.; and Zhang, K. 2022.
\newblock Latent hierarchical causal structure discovery with rank constraints.
\newblock \emph{Advances in Neural Information Processing Systems}, 35: 5549--5561.

\bibitem[{Hyv{\"{a}}rinen, Karhunen, and Oja(2001)}]{aapo2001independent}
Hyv{\"{a}}rinen, A.; Karhunen, J.; and Oja, E. 2001.
\newblock \emph{Independent Component Analysis}.
\newblock Wiley.

\bibitem[{Kummerfeld and Ramsey(2016)}]{kummerfeld2016causal}
Kummerfeld, E.; and Ramsey, J. 2016.
\newblock Causal clustering for 1-factor measurement models.
\newblock In \emph{Proceedings of the 22nd ACM SIGKDD international conference on knowledge discovery and data mining}, 1655--1664.

\bibitem[{Ledermann(1937)}]{Ledermann1937OnTR}
Ledermann, W. 1937.
\newblock On the rank of the reduced correlational matrix in multiple-factor analysis.
\newblock \emph{Psychometrika}, 2: 85--93.

\bibitem[{Lewicki and Sejnowski(2000)}]{lewicki2000learning}
Lewicki, M.~S.; and Sejnowski, T.~J. 2000.
\newblock Learning overcomplete representations.
\newblock \emph{Neural Computation}, 12(2): 337--365.

\bibitem[{Maeda and Shimizu(2020)}]{maeda2020rcd}
Maeda, T.~N.; and Shimizu, S. 2020.
\newblock RCD: Repetitive causal discovery of linear non-Gaussian acyclic models with latent confounders.
\newblock In \emph{International Conference on Artificial Intelligence and Statistics}, 735--745. PMLR.

\bibitem[{Morioka and Hyvarinen(2023)}]{morioka2023connectivity}
Morioka, H.; and Hyvarinen, A. 2023.
\newblock Connectivity-contrastive learning: Combining causal discovery and representation learning for multimodal data.
\newblock In \emph{International Conference on Artificial Intelligence and Statistics}, 3399--3426. PMLR.

\bibitem[{Sachs et~al.(2005)Sachs, Perez, Pe'er, Lauffenburger, and Nolan}]{sachs2005causal}
Sachs, K.; Perez, O.; Pe'er, D.; Lauffenburger, D.~A.; and Nolan, G.~P. 2005.
\newblock Causal protein-signaling networks derived from multiparameter single-cell data.
\newblock \emph{Science}, 308(5721): 523--529.

\bibitem[{Shimizu et~al.(2006)Shimizu, Hoyer, Hyv{\"a}rinen, Kerminen, and Jordan}]{shimizu2006linear}
Shimizu, S.; Hoyer, P.~O.; Hyv{\"a}rinen, A.; Kerminen, A.; and Jordan, M. 2006.
\newblock A linear non-Gaussian acyclic model for causal discovery.
\newblock \emph{Journal of Machine Learning Research}, 7(10).

\bibitem[{Silva et~al.(2006)Silva, Scheines, Glymour, Spirtes, and Chickering}]{silva2006learning}
Silva, R.; Scheines, R.; Glymour, C.; Spirtes, P.; and Chickering, D.~M. 2006.
\newblock Learning the Structure of Linear Latent Variable Models.
\newblock \emph{Journal of Machine Learning Research}, 7(2).

\bibitem[{Spirtes, Meek, and Richardson(1995)}]{spirtes1995causal}
Spirtes, P.; Meek, C.; and Richardson, T. 1995.
\newblock Causal inference in the presence of latent variables and selection bias.
\newblock In \emph{Conference on Uncertainty in artificial intelligence}, 499--506.

\bibitem[{Tashiro et~al.(2014)Tashiro, Shimizu, Hyv{\"a}rinen, and Washio}]{tashiro2014parcelingam}
Tashiro, T.; Shimizu, S.; Hyv{\"a}rinen, A.; and Washio, T. 2014.
\newblock ParceLiNGAM: A causal ordering method robust against latent confounders.
\newblock \emph{Neural computation}, 26(1): 57--83.

\bibitem[{Tramontano, Monod, and Drton(2022)}]{tramontano2022learning}
Tramontano, D.; Monod, A.; and Drton, M. 2022.
\newblock Learning linear non-Gaussian polytree models.
\newblock In \emph{Uncertainty in Artificial Intelligence}, 1960--1969. PMLR.

\bibitem[{Wang and Drton(2020)}]{wang2020high}
Wang, Y.~S.; and Drton, M. 2020.
\newblock High-dimensional causal discovery under non-Gaussianity.
\newblock \emph{Biometrika}, 107(1): 41--59.

\bibitem[{Xie et~al.(2020)Xie, Cai, Huang, Glymour, Hao, and Zhang}]{xie2020generalized}
Xie, F.; Cai, R.; Huang, B.; Glymour, C.; Hao, Z.; and Zhang, K. 2020.
\newblock Generalized independent noise condition for estimating latent variable causal graphs.
\newblock \emph{Advances in neural information processing systems}, 33: 14891--14902.

\bibitem[{Xie et~al.(2022)Xie, Huang, Chen, He, Geng, and Zhang}]{xie2022identification}
Xie, F.; Huang, B.; Chen, Z.; He, Y.; Geng, Z.; and Zhang, K. 2022.
\newblock Identification of linear non-gaussian latent hierarchical structure.
\newblock In \emph{International Conference on Machine Learning}, 24370--24387. PMLR.

\end{thebibliography}
\newpage
\appendix
\onecolumn

\section*{Appendix}

In App.A, we provide the proof of theorems. In App.B, we provide more experimental results. 

\section{Proofs}
\subsection{Proof of Theorem 4.1}
\textbf{Theorem 4.1} Assume that two observed variables $X$ and $Y$ are generated by Eq. (\ref{eq:model}). Suppose there exists $i, j > 0$, such that $\mathcal{C}_{i+3, j}(X, Y)$, $\mathcal{C}_{i+2, j+1}(X, Y)$, $\mathcal{C}_{i+1, j+2}(X, Y)$ and $\mathcal{C}_{i, j+3}(X, Y)$ are not equal to zero. If $X$ and $Y$ are affected by two independent components with different ratios of mixing coefficients, then the ratio of mixing coefficients $\theta_1$ and $\theta_2$ of two independent components on $X$ and $Y$ can be estimated by the analytical solution of the following equation:
\begin{equation}
\begin{aligned}
    &\left( \mathcal{C}_{i+2, j+1}(X, Y)^2 - \mathcal{C}_{i+1, j+2}(X, Y) \mathcal{C}_{i+3, j}(X, Y) \right) \theta^{2} \\
    + & \mathcal{C}_{i, j+3}(X, Y)\mathcal{C}_{i+3, j}(X, Y) \theta \\
    -& \mathcal{C}_{i+1, i+2}(X, Y) \mathcal{C}_{i+2, j+1}(X, Y) \theta\\
    + & \mathcal{C}_{i+1, j+2}(X, Y)^2 - \mathcal{C}_{i, j+3}(X, Y) \mathcal{C}_{i+2, j+1}(X, Y) \\
    = & 0.
\end{aligned}
\end{equation}

\begin{proof}
Suppose $X$ and $Y$ are affected by two independent components $S_1$ and $S_2$. Then, the generating process of $X$ and $Y$ can be written in terms of the mixing matrix:
\begin{equation}
    \begin{aligned}
        \left[\begin{array}{cc}
         X   \\
         Y 
        \end{array}\right]=
         \begin{bmatrix}
        \alpha _{1} & \beta _{1} & \mathbf{A}_{X, \mathbf{S'}_X} & \mathbf{0}\\
        \alpha _{2} & \beta _{2} & \mathbf{0} & \mathbf{A}_{Y, \mathbf{S'}_Y}
        \end{bmatrix}  \begin{bmatrix}
        S_{1}\\
        S_{2}\\
        \mathbf{S'}_X \\
        \mathbf{S'}_Y
        \end{bmatrix},
    \end{aligned}
    \label{eq:proof_1_generation}
\end{equation}
where $\alpha_1$ and $\alpha_2$ are the mixing coefficients of $S_1$ on $X$ and $Y$, respectively. $\beta_1$ and $\beta_2$ are the mixing coefficients of $S_2$ on $X$ and $Y$, respectively. $S_1$, $S_2$, $\mathbf{S'}_X$ and $\mathbf{S'}_Y$ are mutually independent components. 

Let $\theta_1 = \frac{\alpha_2}{\alpha_1}$ and $\theta_2 = \frac{\beta_2}{\beta_1}$. Then, there exist two cases that need to be considered: 1) $\theta_1 \neq \theta_2 \neq 0$; 2) $\theta_1 = \theta_2$, $\theta_1 \neq 0$ and $\theta_2 \neq 0$.

1) Considering the case that $\theta_1 \neq \theta_2 \neq 0$, we utilize the cumulants $\mathcal{C}_{i+3, j}(X, Y)$, $\mathcal{C}_{i+2, j+1}(X, Y)$, $\mathcal{C}_{i+1, j+2}(X, Y)$ and $\mathcal{C}_{i, j+3}(X, Y)$ (where $i > 0$ and $j > 0$):
\begin{equation}
    \begin{aligned}
    \mathcal{C}_{i+3, j}(X, Y) & = \alpha_{1}^{i+3}\alpha_{2}^{j} \mathcal{C}_{i+j+3}(S_1) + \beta_{1}^{i+3}\beta_{2}^{j} \mathcal{C}_{i+j+3}(S_2),\\
    \mathcal{C}_{i+2, j+1}(X, Y) & = \alpha_{1}^{i+2}\alpha_{2}^{j+1} \mathcal{C}_{i+j+3}(S_1) + \beta_{1}^{i+2}\beta_{2}^{j+1} \mathcal{C}_{i+j+3}(S_2),\\
    \mathcal{C}_{i+1, j+2}(X, Y) & = \alpha_{1}^{i+1}\alpha_{2}^{j+2} \mathcal{C}_{i+j+3}(S_1) + \beta_{1}^{i+1}\beta_{2}^{j+2} \mathcal{C}_{i+j+3}(S_2),\\
    \mathcal{C}_{i, j+3}(X, Y) & = \alpha_{1}^{i}\alpha_{2}^{j+3} \mathcal{C}_{i+j+3}(S_1) + \beta_{1}^{i}\beta_{2}^{j+3} \mathcal{C}_{i+j+3}(S_2).\\
    \end{aligned}
\end{equation}
Then we can obtain: 
\begin{equation}
    \begin{aligned}
    &\frac{\alpha_2}{\alpha_1}\mathcal{C}_{i+3, j}(X, Y) - \mathcal{C}_{i+2, j+1}(X, Y) = \left(\frac{\alpha_2}{\alpha_1} -  \frac{\beta_2}{\beta_1}\right) \beta_{1}^{i+3}\beta_{2}^{j} \mathcal{C}_{i+j+3}(S_2),\\
    \end{aligned}
    \label{eq:th_eq1}
\end{equation}
\begin{equation}
    \begin{aligned}
     &\frac{\alpha_2}{\alpha_1}\mathcal{C}_{i+2, j+1}(X, Y) - \mathcal{C}_{i+1, j+2}(X, Y) = \left(\frac{\alpha_2}{\alpha_1} -  \frac{\beta_2}{\beta_1}\right) \beta_{1}^{i+2}\beta_{2}^{j+1} \mathcal{C}_{i+j+3}(S_2),\\
    \end{aligned}
    \label{eq:th_eq2}
\end{equation}
and 
\begin{equation}
    \begin{aligned}
     \frac{\alpha_2}{\alpha_1}\mathcal{C}_{i+1, j+2}(X, Y) - \mathcal{C}_{i, j+3}(X, Y) = \left(\frac{\alpha_2}{\alpha_1} -  \frac{\beta_2}{\beta_1}\right) \beta_{1}^{i+1}\beta_{2}^{j+2} \mathcal{C}_{i+j+3}(S_2).\\
    \end{aligned}
    \label{eq:th_eq3}
\end{equation}
Combing Eq. (\ref{eq:th_eq1}) - (\ref{eq:th_eq3}), we have:
\begin{equation}
\begin{aligned}
    \frac{\frac{\alpha_2}{\alpha_1}\mathcal{C}_{i+1, j+2}(X, Y) - \mathcal{C}_{i, j+3}(X, Y)}{\frac{\alpha_2}{\alpha_1}\mathcal{C}_{i+2, j+1}(X, Y) - \mathcal{C}_{i+1, j+2}(X, Y)} = \frac{\beta_2}{\beta_1}.
\end{aligned}
\end{equation}
Let $\theta = \frac{\alpha_2}{\alpha_1}$, we can obtain the quadratic equation as follows:
\begin{equation}
\begin{aligned}
&\left( \mathcal{C}_{i+2, j+1}(X, Y)^2 - \mathcal{C}_{i+3, j}(X, Y)\mathcal{C}_{i+1, j+2}(X, Y) \right)\theta^2 \\
+ & \left(\mathcal{C}_{i, j+3}(X, Y)\mathcal{C}_{i+3, j}(X, Y) - \mathcal{C}_{i+1, j+2}(X, Y)\mathcal{C}_{i+2, j+1}(X, Y) \right) \theta\\
+ & \mathcal{C}_{i+1, j+2}(X, Y)^2 -   \mathcal{C}_{i, j+3}(X, Y)\mathcal{C}_{i+2, j+1}(X, Y)\\
= & 0.\\
\end{aligned}
\end{equation}

For the ratio $\theta = \frac{\beta_2}{\beta_1}$, we can obtain the same quadratic equation as above. Thus, the two solutions of the quadratic equation (\ref{eq:general_quadratic_equation}) are the ratios $\frac{\alpha_2}{\alpha_1}$ and $\frac{\beta_2}{\beta_1}$.

2) Considering the case that $\theta_1 = \theta_2$, $\theta_1 \neq 0$ and $\theta_2 \neq 0$,  i.e., $\frac{\alpha_2}{\alpha_1} = \frac{\beta_2}{\beta_1}$. The data generation process can be rewritten as:

\begin{equation}
    \begin{aligned}
        X &= \zeta_X S_s + \mathbf{A}_{X, \mathbf{S'}_X}\mathbf{S'}_X,\\
        Y &= \zeta_Y S_s + \mathbf{A}_{Y, \mathbf{S'}_Y}\mathbf{S'}_Y,
    \end{aligned}
\end{equation}
where $\zeta_X S_s = \alpha_1 S_1 + \beta_1 S_2$, $\zeta_Y S_s = \alpha_2 S_1 + \beta_2 S_2$. 

Then, the cumulants $\mathcal{C}_{i+3, j}(X, Y)$, $\mathcal{C}_{i+2, j+1}(X, Y)$, $\mathcal{C}_{i+1, j+2}(X, Y)$ and $\mathcal{C}_{i, j+3}(X, Y)$ can be obtained as follows:
\begin{equation}
    \begin{aligned}
    \mathcal{C}_{i+3, j}(X, Y) & = \zeta_X^{i+3}\zeta_Y^{j}\mathcal{C}_{i+j+3}(S_s),\\
    \mathcal{C}_{i+2, j+1}(X, Y) & = \zeta_X^{i+2}\zeta_Y^{j+1}\mathcal{C}_{i+j+3}(S_s),\\
    \mathcal{C}_{i+1, j+2}(X, Y) & = \zeta_X^{i+1}\zeta_Y^{j+2}\mathcal{C}_{i+j+3}(S_s),\\
    \mathcal{C}_{i, j+3}(X, Y) & = \zeta_X^{i}\zeta_Y^{j+3}\mathcal{C}_{i+j+3}(S_s).\\
    \end{aligned}
\end{equation}

Thus, 
\begin{equation}
\begin{aligned}
\left( \mathcal{C}_{i+2, j+1}(X, Y)^2 - \mathcal{C}_{i+3, j}(X, Y)\mathcal{C}_{i+1, j+2}(X, Y) \right)\theta^2 &= 0,\\
\left(\mathcal{C}_{i, j+3}(X, Y)\mathcal{C}_{i+3, j}(X, Y) -\mathcal{C}_{i+1, i+2}(X, Y) \mathcal{C}_{i+2, j+1}(X, Y)\right) &= 0,\\
\mathcal{C}_{i+1, j+2}(X, Y)^2 - \mathcal{C}_{i, j+3}(X, Y) \mathcal{C}_{i+2, j+1}(X, Y) & = 0,
\end{aligned}
\end{equation}
which makes the quadratic equation of $\theta$ have infinite solutions. So it can not estimate the ratio of mixing coefficients.

Therefore, the theorem is proven.
\end{proof}

\subsection{Proof of Theorem 4.2}

\textbf{Theorem 4.2}
Assume that two observed variables $X$ and $Y$ are generated by Eq. (\ref{eq:model}), and $X$ and $Y$ are affected by the same latent variable $L$. Suppose there exists $i, j > 0$, such that $\mathcal{C}_{i, j+2}(X, Y)$, $\mathcal{C}_{i+1, j+1}(X, Y)$ and $\mathcal{C}_{i+2, j}(X, Y)$ are not equal to zero. Then there is no directed edge between $X$ and $Y$, if and only if $\mathcal{C}_{i+1, j+1}(X, Y)^2 - \mathcal{C}_{i, j+2}(X, Y) \mathcal{C}_{i+2, j}(X, Y) = 0$.   
\label{th: edge_existence}

\begin{proof}
    Assume that the data over two observed variables $X$ and $Y$ are generated by Eq. (\ref{eq:model}). Then, we can obtain the following equations:
    \begin{equation}
        \begin{aligned}
            X &= \alpha_1 L + \beta_1 E_X + \gamma_1 E_Y,\\
            Y &= \alpha_2 L + \beta_2 E_X + \gamma_2 E_Y.        
        \end{aligned}
        \label{eq: edge_existence_model}
    \end{equation}
    1) The ``if" part: According to Eq. (\ref{eq: edge_existence_model}) and the model assumptions, we have $\beta_1\beta_2 = 0$ or $\gamma_1\gamma_2 = 0$. Then, for $i,j>0$, the higher-order joint cumulants of $X$ and $Y$ are
    \begin{equation}
        \begin{aligned}
            \mathcal{C}_{i+1, j+1}(X, Y) &= \alpha_{1}^{i+1} \alpha _{2}^{j+1} \mathcal{C}_{i+j+2}(L) +\beta_{1}^{i+1} \beta_{2}^{j+1} \mathcal{C}_{i+j+2}(E_{X})+\gamma_{1}^{i+1} \gamma_{2}^{j+1} \mathcal{C}_{i+j+2}(E_{Y}),\\
            \mathcal{C}_{i, j+2}(X, Y) &= \alpha_{1}^{i} \alpha _{2}^{j+2} \mathcal{C}_{i+j+2}(L) +\beta_{1}^{i} \beta_{2}^{j+2} \mathcal{C}_{i+j+2}(E_{X})+\gamma_{1}^{i} \gamma_{2}^{j+2} \mathcal{C}_{i+j+2}(E_{Y}),\\
            \mathcal{C}_{i+2, j}(X, Y) &=\alpha_{1}^{i+2} \alpha _{2}^{j} \mathcal{C}_{i+j+2}(L) +\beta_{1}^{i+2} \beta_{2}^{j} \mathcal{C}_{i+j+2}(E_{X})+\gamma_{1}^{i+2} \gamma_{2}^{j} \mathcal{C}_{i+j+2}(E_{Y}).
        \end{aligned}
    \end{equation}

    If $\mathcal{C}_{i+1, j+1}(X, Y)^2 - \mathcal{C}_{i, j+2}(X, Y) \mathcal{C}_{i+2, j}(X, Y) = 0$, then 
    \begin{equation}
        \begin{aligned}
            &\mathcal{C}_{i+1, j+1}(X, Y)^2 - \mathcal{C}_{i, j+2}(X, Y) \mathcal{C}_{i+2, j}(X, Y)\\
            = & \alpha _{1}^{i} \alpha _{2}^{j} \beta _{1}^{i} \beta _{2}^{j}\left( 2\alpha _{1} \alpha _{2} \beta _{1} \beta _{2} -\alpha _{2}^{2} \beta _{1}^{2} -\alpha _{1}^{2} \beta _{2}^{2}\right) \mathcal{C}_{i+j+2} (L)\mathcal{C}_{i+j+2} (E_{X})\\
            &+  \alpha _{1}^{i} \alpha _{2}^{j}\gamma _{1}^{i} \gamma _{2}^{j}\left( 2\alpha _{1} \alpha _{2} \gamma _{1} \gamma _{2} -\alpha_{2}^{2}\gamma _{1}^{2} -\alpha _{1}^{2} \gamma _{2}^{2}\right)\mathcal{C}_{i+j+2} (L)\mathcal{C}_{i+j+2} (E_{Y} )\\
             &+ \beta _{1}^{i} \beta _{2}^{j}\gamma _{1}^{i} \gamma _{2}^{j}\left( 2\beta _{1} \beta _{2} \gamma _{1} \gamma _{2} -\beta _{2}^{2}\gamma _{1}^{2} -\beta _{1}^{2} \gamma _{2}^{2}\right)\mathcal{C}_{i+j+2} ( E_{X})\mathcal{C}_{i+j+2} (E_{Y} )\\
            =&- \alpha _{1}^{i} \alpha _{2}^{j} \beta _{1}^{i} \beta _{2}^{j}\left(\alpha_1\beta_2 - \alpha_2\beta_1\right)^2 \mathcal{C}_{i+j+2} (L)\mathcal{C}_{i+j+2} (E_{X})\\ 
            &- \alpha _{1}^{i} \alpha _{2}^{j} \gamma _{1}^{i} \gamma_{2}^{j}\left(\alpha_1\gamma_2 - \alpha_2\gamma_1\right)^2 \mathcal{C}_{i+j+2} (L)\mathcal{C}_{i+j+2} (E_{X})\\ 
            &- \beta _{1}^{i} \beta _{2}^{j} \gamma _{1}^{i} \gamma_{2}^{j}\left(\beta_1\gamma_2 - \beta_2\gamma_1\right)^2 \mathcal{C}_{i+j+2} (L)\mathcal{C}_{i+j+2} (E_{X})\\
            &= 0.\\
            \end{aligned}
    \end{equation}
Note that the quadratic terms (e.g., $\left(\alpha_1\beta_2 - \alpha_2\beta_1\right)^2$) must be greater than or equal to 0. To make the above equation hold, it must satisfy a condition that there are at least two terms to be zero among $\alpha_{1}^{i} \alpha_{2}^{j}$, $\beta_{1}^{i} \beta_{2}^{j}$ and $\gamma _{1}^{i} \gamma_{2}^{j}$, which indicate how many latent components they are affecting $X$ and $Y$. That is, $X$ and $Y$ only share one latent component. This one latent component should be the latent confounder accroding to the assumption. Thus, there is no directed edge between $X$ and $Y$.


2) The ``only if" part: suppose the data over variables are generated according to a graph that there is no directed edge between $X$ and $Y$, and they are affected by a latent confounder. Then, we can obtain the following equations:
    \begin{equation}
        \begin{aligned}
            X &= \alpha_1 L + \beta_1 E_X,\\
            Y &= \alpha_2 L  + \gamma_2 E_Y.        
        \end{aligned}
        \label{eq: edge_existence_graph}
    \end{equation}

According to Eq. (\ref{eq: edge_existence_graph}), 
 \begin{equation}
        \begin{aligned}
            \mathcal{C}_{i+1, j+1}(X, Y) &= \alpha_{1}^{i+1} \alpha _{2}^{j+1} \mathcal{C}_{i+j+2}(L),\\
            \mathcal{C}_{i, j+2}(X, Y) &= \alpha_{1}^{i} \alpha _{2}^{j+2} \mathcal{C}_{i+j+2}(L),\\
            \mathcal{C}_{i+2, j}(X, Y) &=\alpha_{1}^{i+2} \alpha _{2}^{j} \mathcal{C}_{i+j+2}(L).
        \end{aligned}
    \end{equation}
    
Then, 
\begin{equation}
    \mathcal{C}_{i+1, j+1} (X,Y)^{2} -\mathcal{C}_{i,j+2} (X,Y)\mathcal{C}_{i+2,j} (X,Y)= 0.
\end{equation}
    Thus, the ``only if" part is proven.

    From the above proof, the theorem holds.
\end{proof}

\subsection{Proof of Theorem 4.3}
\textbf{Theorem 4.3}
Assume that two observed variables $X$ and $Y$ are generated by Eq. (\ref{eq:model}), and $X$ and $Y$ are affected by the same latent variable $L$. Then $X$ is a cause of $Y$ if and only if $\mathcal{R}_{X \to Y} = 0$.

\begin{proof}
Assume that the data over two observed variables $X$ and $Y$ are generated by Eq. (\ref{eq:model}), then the data generating process of $X$ and $Y$ can be formalized as:
\begin{equation}
    \begin{aligned}
            X &= \alpha_1 L + \beta_1 E_X + \gamma_1 E_Y,\\
            Y &= \alpha_2 L + \beta_2 E_X + \gamma_2 E_Y.        
        \end{aligned}
\end{equation}

     1) The ``if" part: According to Eq. (\ref{eq: edge_existence_model}) and the model assumptions, we have $\beta_1\beta_2 = 0$ or $\gamma_1\gamma_2 = 0$. Then,
     \begin{equation}
        \begin{aligned}
            \mathcal{R}_{X\rightarrow Y}  & = \mathcal{C}_{3} (X)-\alpha _{1}^{3}\mathcal{C}_{3} (L)-\beta _{1}^{3}\mathcal{C}_{3} (E_{X})\\
            & = \alpha _{1}^{3}\mathcal{C}_{3} (L)+\beta _{1}^{3}\mathcal{C}_{3} (E_{X})+ \gamma_1^3 \mathcal{C}_{3} (E_{Y}) - \alpha _{1}^{3}\mathcal{C}_{3} (L)-\beta _{1}^{3}\mathcal{C}_{3} (E_{X}),\\
            & = \gamma_1^3 \mathcal{C}_{3} (E_{Y}).
        \end{aligned}
     \end{equation}

Because $\mathcal{C}_{3} (E_{Y}) \neq 0$, $\gamma_1$ should be zero if $\mathcal{R}_{X \to Y} = 0$. Then $\beta_1\beta_2 \neq 0$ due to $\gamma_1\gamma_2 = 0$. That is, $X$ is a cause of $Y$.

    2) The ``only if" part: suppose $X$ is a cause of $Y$, then the data generating process of $X$ and $Y$ can be formalized as:
    \begin{equation}
        \begin{aligned}
            X & = \lambda_1 L + \beta_1 E_X,\\
            Y & = b_{2} X + \lambda_{2} L + \gamma_2 E_Y \\
              & = (b_2\lambda_1 + \lambda_2) L +b_2\beta_1 E_X + \gamma_2 E_Y,
        \end{aligned}
    \end{equation}
    where $b_{2}$ denotes the causal coefficient of $X$ on $Y$, $\lambda_1$ and $\lambda_2$ denotes the causal coefficients of $L$ on $X$ and $Y$, respectively. 

    Then, it is easily to obtain
    \begin{equation}
\mathcal{R}_{X\rightarrow Y} = \mathcal{C}_{3} (X)-\lambda _{1}^{3}\mathcal{C}_{3} (L)-\beta _{1}^{3}\mathcal{C}_{3} (E_{X}) =0.
    \end{equation}

 From the above analysis, the theorem holds.
\end{proof}

\subsection{Proof of Theorem 4.4}
\textbf{Theorem 4.4}
Assume that two observed variables $X$ and $Y$ are generated by Eq. (\ref{eq:model}), and $X$ and $Y$ are affected by the same latent variable $L$. The causal relationship between two observed variables can be identified by using the higher-order cumulants.  

\begin{proof}
    Assume that two observed variables $X$ and $Y$ are generated by Eq. (\ref{eq:model}), and $X$ and $Y$ are affected by the same latent variable $L$. Then the causal structure between $X$ and $Y$ is one of the following three kinds:
    \begin{enumerate}
        \item There is no directed edge between $X$ and $Y$;
        \item $X$ is a cause of $Y$;
        \item $X$ is a effect of $Y$.
    \end{enumerate}

    The difference between the first case and the last two cases is the existence of the directed edge between $X$ and $Y$. This difference can be identified by using Theorem 4.2, which leverages the higher-order cumulants. To identify the last two cases, we can utilize Theorem 4.3 to achieve this by using higher-order cumulants.

    Therefore, the Theorem holds.
\end{proof}
\section{More experimental results}

We conduct an experiment under nonlinear generation process, the detail results are shown in the Table \ref{tab:nonlinear-no-edge} and Table \ref{tab:nonlinear-case2}. The nonlinear causal mechanism can be formalized as follows: 
\begin{equation}
    V_i = \sum_{j} (B_{i, j}V_j + D_{i, j} V_{j}^3)+ N_i, 
\end{equation}
where $B_{i, j}$ is the coefficient of linear terms, $D_{i, j}$ is the coefficient of nonlinear terms. The coefficient of nonlinear terms is sample from a uniform distribution between $[0.01, 0.03]$.

We can observe that our method still performs well expect in the gamma distribution. We found that the standard deviation of this gamma random variable will increase by more than 60 times after being transformed by a cubic function. For the random variables of the other two distributions, the corresponding standard deviation will only increase by 20 to 30 times. In the gamma distribution, the nonlinear term will be more important than the other distribution, so it will perform bad than the others. It also shows that our method can be applied in the scenario with slightly nonlinearity. 

\begin{table}[htbp]
  \centering
    \begin{tabular}{cccccc}
    \toprule
          & Sample Size & LiNGAM & ANM   & LvLiNGAM & Ours \\
    \midrule
    \multirow{4}[2]{*}{Exp} & 5000  & 0.00  & -     & 0.00  &\textbf{ 0.56}  \\
          & 10000 & 0.00  & -     & 0.00  & \textbf{0.61}  \\
          & 50000 & 0.00  & -     & 0.00  & \textbf{0.60}  \\
          & 100000 & 0.00  & -     & 0.00  & \textbf{0.66}  \\
    \midrule
    \multirow{4}[2]{*}{Gamma} & 5000  & 0.00  & -     & 0.00  & \textbf{0.64}  \\
          & 10000 & 0.00  & -     & 0.00  & \textbf{0.55}  \\
          & 50000 & 0.00  & -     & 0.00  & \textbf{0.69}  \\
          & 100000 & 0.00  & -     & 0.00  & \textbf{0.66}  \\
    \midrule
    \multirow{4}[2]{*}{Gumbel} & 5000  & 0.00  & -     & 0.00  & \textbf{0.53}  \\
          & 10000 & 0.00  & -     & 0.00  & \textbf{0.60}  \\
          & 50000 & 0.00  & -     & 0.00  & \textbf{0.61}  \\
          & 100000 & 0.00  & -     & 0.00  & \textbf{0.75}  \\
    \bottomrule
    \end{tabular}%
    \caption{Accuracy of the different methods varies across different distributions in Case 1 under nonlinear generation process. Here ``-" denotes that the corresponding methods are not applicable.}
  \label{tab:nonlinear-no-edge}%
\end{table}%

\begin{table}[htbp]
  \centering
    \begin{tabular}{cccccc}
    \toprule
          & Sample Size & LiNGAM & ANM   & LvLiNGAM & Ours \\
    \midrule
    \multirow{4}[2]{*}{Exp} & 5000  & 0.02  & 0.33  & 0.50  & \textbf{0.62}  \\
          & 10000 & 0.02  & 0.08  & 0.58  & \textbf{0.64}  \\
          & 50000 & 0.04  & -     & 0.59  & \textbf{0.77}  \\
          & 100000 & 0.03  & -     & 0.56  & \textbf{0.84}  \\
    \midrule
    \multirow{4}[2]{*}{Gamma} & 5000  & 0.00  & 0.28  & \textbf{0.45}  & 0.38  \\
          & 10000 & 0.00  & 0.07  & \textbf{0.46}  & 0.36  \\
          & 50000 & 0.01  & -     & \textbf{0.43}  & 0.34  \\
          & 100000 & 0.04  & -     & \textbf{0.54}  & 0.37  \\
    \midrule
    \multirow{4}[2]{*}{Gumbel} & 5000  & 0.06  & 0.48  & \textbf{0.53}  & 0.44  \\
          & 10000 & 0.06  & 0.34  & 0.54  & \textbf{0.68}  \\
          & 50000 & 0.00  & -     & 0.49  & \textbf{0.68 } \\
          & 100000 & 0.03  & -     & 0.52  & \textbf{0.72}  \\
    \bottomrule
    \end{tabular}%
    \caption{Accuracy of the different methods varies across different distributions in Case 2 under nonlinear generation process. Here ``-" denotes that the corresponding methods take too long to produce any results.}
  \label{tab:nonlinear-case2}
\end{table}

\end{document}